\documentclass{article}
\usepackage{iclr2027_conference,times}

\usepackage{amsmath,amsfonts,bm}

\def\eqref#1{equation~\ref{#1}}

\def\1{\bm{1}}

\DeclareMathAlphabet{\mathsfit}{\encodingdefault}{\sfdefault}{m}{sl}
\SetMathAlphabet{\mathsfit}{bold}{\encodingdefault}{\sfdefault}{bx}{n}

\usepackage{hyperref,url,graphicx,booktabs,array,tabularx,amsmath,amssymb,tikz}
\usepackage{xurl}
\usepackage{wrapfig,needspace}
\usetikzlibrary{arrows.meta,positioning}
\hypersetup{hidelinks}
\newcommand{\nexteval}{NextEval}
\newcommand{\code}[1]{\texttt{#1}}
\title{\texorpdfstring{\raisebox{-0.08em}{\includegraphics[height=0.82em]{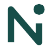}}\hspace{0.3em}}{}
NextEval: An Agent-Based Solver and\\
Benchmark for Derivative-Free Optimization}
\iclrfinalcopy
\author{%
\textnormal{Shuo Chen\textsuperscript{1,$\dagger$} \quad Li Cui\textsuperscript{2} \quad Cunxin Huang\textsuperscript{3,$\dagger$,*} \quad Ya-Feng Liu\textsuperscript{4}}\\
\textnormal{Minghang Xie\textsuperscript{2,$\dagger$} \quad Meng Xu\textsuperscript{5,$\dagger$,*} \quad Zaikun Zhang\textsuperscript{6}}\\[4pt]
{\normalfont\footnotesize
\begin{tabular}{@{}l@{\hspace{1.1em}}l@{}}
\textsuperscript{1}Independent Researcher & \textsuperscript{2}Beijing Normal University\\
\textsuperscript{3}The Hong Kong Polytechnic University & \textsuperscript{4}Beijing University of Posts and Telecommunications\\
\textsuperscript{5}University of the Chinese Academy of Sciences & \textsuperscript{6}Sun Yat-sen University
\end{tabular}}\\[3pt]
\textnormal{\footnotesize Authors are listed in alphabetical order by surname.\quad $\dagger$ Project leads.\quad * Corresponding authors.}%
}
\hypersetup{
  pdftitle={NextEval: An Agent-Based Solver and Benchmark for Derivative-Free Optimization},
  pdfauthor={Shuo Chen, Li Cui, Cunxin Huang, Ya-Feng Liu, Minghang Xie, Meng Xu, Zaikun Zhang}
}
\begin{document}
\maketitle
\lhead{Preprint}
\begin{abstract}
Derivative-free optimization (DFO) addresses problems where derivatives
are unavailable or costly to obtain. This setting often arises in
simulation-based and experimental applications, where evaluating the black-box objective can itself be costly. Thus, it remains a central challenge in DFO to effectively exploit accumulated observations to select informative evaluation points under a limited budget. In this light, we introduce the NextEval solver, an agent-based DFO solver, which integrates the capabilities of a large language model (LLM) to guide the sequential selection of evaluation points. Specifically, we establish a system that combines an LLM agent with a numerical environment. The environment provides tools for analyzing past evaluations and executing numerical methods, while maintaining a shared evaluation history and method-specific state to support continuation across method switches. Using observations and tool feedback, the LLM agent selects tools, configures their parameters, and allocates the evaluation budget, while numerical methods compute and evaluate candidate points.
Building on this system, we also introduce NextEval Bench, a benchmark protocol for comparing agent configurations on shared DFO tasks using the same numerical tools under matched evaluation budgets and information constraints.
Numerical results on low-dimensional DFO problems demonstrate that the NextEval solver outperforms the state-of-the-art DFO baselines in function-evaluation efficiency.
\end{abstract}

\section{Introduction}
\label{sec:intro}
\paragraph{Derivative-free optimization.}
Derivative-free optimization (DFO) addresses optimization problems in
which derivatives are unavailable, unreliable, or too expensive to
compute \citep{conn2009idfo,larson2019dfo}. Such problems arise, for
example, when objectives are evaluated through simulations or experiments
\citep{audet2017dfbook}. These evaluations are often expensive, so a
central goal is to obtain a good solution within a limited evaluation budget
\citep{morewild2009benchmarking}. Therefore, DFO algorithms are commonly compared
by the number of objective evaluations needed to reach a given quality, not by runtime.

\paragraph{Choosing and combining numerical methods.}
Classical DFO chooses evaluation points through direct search or local
approximations \citep{kolda2003,powell2004frobenius,powell2006newuoa}.
As function evaluations accumulate, the algorithm decides whether to continue,
change parameters, improve the sample set, or try another numerical method.
These choices can depend on progress, the geometry of the sample set, and
agreement between local-model predictions and actual function changes.
\textbf{We ask whether these choices can be made by a language-model agent
instead of predefined strategies coded in the algorithm.} We address this
question in unconstrained, low-dimensional DFO with a small evaluation budget.

\paragraph{Agent-guided search.}
Large language model (LLM) agents can use tools and respond to their outputs
over multiple interactions \citep{yao2023react}; agent frameworks support
conversation state and repeated LLM--tool interactions
\citep{langgraph2026overview}. The agent uses evaluation histories to select
numerical methods, while numerical tools compute the candidate points.

\paragraph{A sequential policy, not a solution mapping.}
In our study, the agent pursues a history-dependent policy, not a
\emph{solution mapping} from a fixed input to a minimizer: new evaluations can change the next operation
and its parameters.
Unlike trained DFO policies \citep{rw20260906_chen_learned_dfo}, our agent uses
general-purpose LLMs without optimization-specific policy training.

\begin{figure}[t]
\centering
\includegraphics[width=.90\linewidth,trim={0bp 32bp 0bp 70bp},clip]{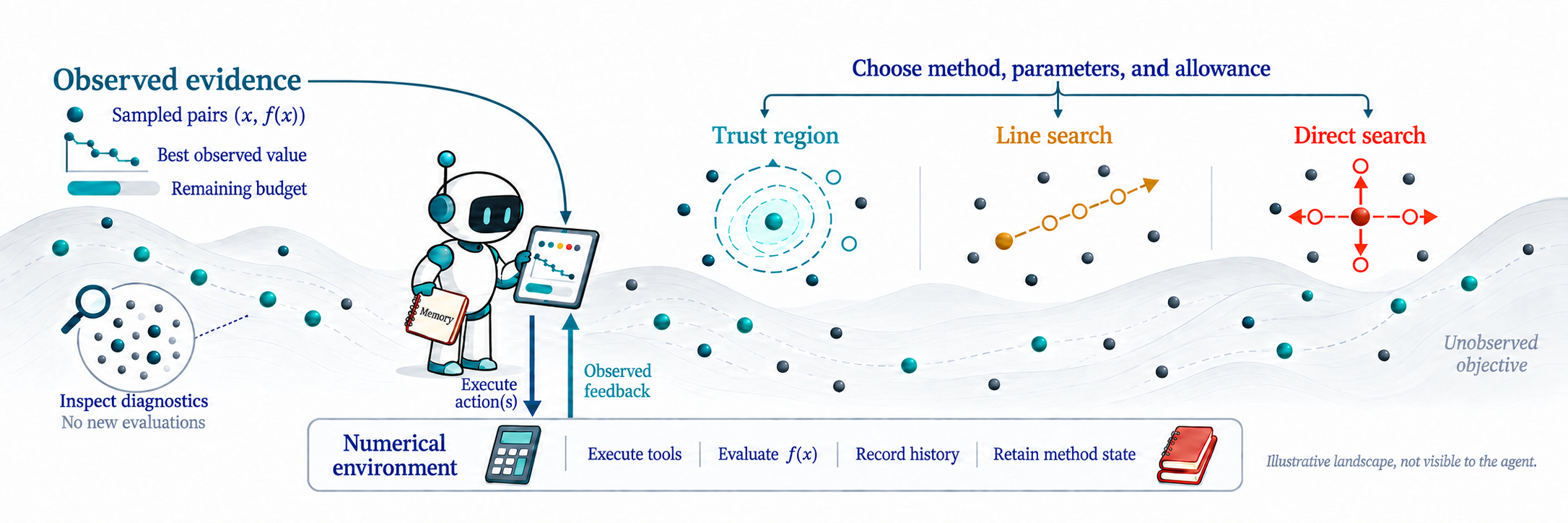}
\caption{Conceptual view of NextEval. The agent selects tools using past
evaluations. The illustrated landscape is neither an agent input nor an
experimental trajectory.}
\label{fig:concept}
\end{figure}

We introduce \nexteval{}, an agent-based DFO solver for unconstrained
minimization. Its numerical environment executes the agent's tool requests,
records objective calls, and retains method state while enforcing budget
and stopping rules (Figure~\ref{fig:concept}).

NextEval Bench reuses this environment to compare LLMs and harnesses, the
software that manages LLM calls, conversation history, and tool exchanges.

\paragraph{Contributions.}
\begin{itemize}
\item \textbf{An agent-based DFO solver.} NextEval combines a persistent
agent with a numerical environment: executable tools, shared evaluation
history, and method-specific continuation state. Under a shared evaluation
budget, the agent combines single-step and multi-step operations, choosing
their sequence, parameters, and evaluation allowances.
\item \textbf{A benchmark protocol for optimization agents.} NextEval
Bench runs each LLM--harness configuration on individual optimization problems
in a common environment and summarizes its performance across the problem set.
It builds on OptiProfiler, an optimization benchmarking platform with diverse
problem libraries, configurable test conditions, and reproducible performance
and data profiles \citep{optiprofiler}.
\end{itemize}
On the low-dimensional S2MPJ suite, NextEval has higher aggregate
evaluation-efficiency scores than the tested DFO baselines. We also compare
function-evaluation efficiency across LLMs, reasoning settings, and agent
harnesses, using matched tasks and evaluation budgets within each study.

\paragraph{Objective access and the evaluation budget.}
\label{sec:problem}
We consider unconstrained minimization
\begin{equation}
 \min_{x\in\mathbb R^n} f(x),\qquad x_0\in\mathbb R^n,
 \qquad N_{\mathrm{eval}}\le B.
 \label{eq:problem}
\end{equation}
Here, $N_{\mathrm{eval}}$ counts objective evaluations and $B$ is the evaluation budget.
The solver receives a callable \code{fun}, an initial point \code{x0},
and options. It does not receive the objective formula, benchmark
identifier, derivatives, or a reference optimum.
The agent sees only the permitted evaluation history and summaries computed
from it,
and generic metadata such as dimension and remaining budget.

Only objective calls consume this budget: analyzing existing observations
requires no new evaluations, whereas a search tool may make several
calls before returning control.

\section{NextEval Solver}
\label{sec:solver}

\begin{figure}[t]
\centering
\includegraphics[width=\linewidth]{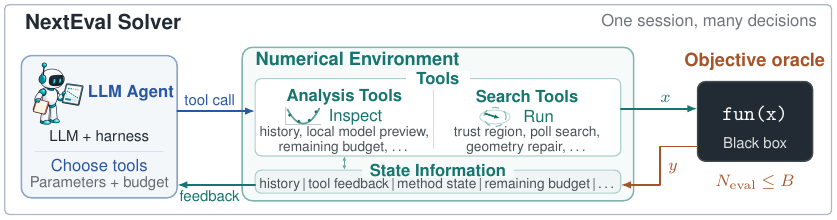}
\caption{Overview of NextEval Solver: the agent selects tools and their parameters, while the numerical environment executes the requests, maintains the evaluation history and the numerical state of each method, and returns feedback. }
\label{fig:architecture}
\end{figure}

In this section, we establish the NextEval solver, as shown in Figure \ref{fig:architecture}, which combines an LLM agent
with a numerical environment for solving DFO problems.
The central idea is to let the agent reason about the search using accumulated
observations, while numerical methods compute and evaluate candidate points.
Here, \emph{numerical methods} refer to components such as model fitting or
trust-region steps, rather than complete DFO algorithms. A \emph{tool} is the
interface through which the agent invokes such a component.
We provide analysis and search tools, and maintain numerical state so that
methods can resume after tool switches.

\subsection{Agent--Environment Interaction}
\label{sec:solver-objective-access}
The agent and the numerical environment interact through tool calls: the agent
selects a tool and its parameters, and the environment executes the request
and returns feedback. This feedback allows the agent to reconsider its search
strategy as new observations become available. In particular, the agent can decide whether to continue the current method, adjust its parameters, or switch to another method.
Specifically, the numerical environment manages all objective evaluations, maintains a shared evaluation history and budget, and retains each method's numerical state. It also maintains the \emph{incumbent}, the evaluated point with the smallest valid observed value across all search tools. This shared record allows the agent to assess progress across methods and base later decisions on observations collected by any of them. To connect the agent to these tools, a \emph{harness} manages the conversation, handles LLM calls, and delivers tool requests and responses. In this setup, the agent decides how to search, the numerical environment performs the numerical work, and the harness coordinates their interaction.

Formally, let \(x_k\) denote the point queried in the \(k\)th objective evaluation and let \(H_k=((x_i,y_i,s_i))_{i=1}^{k}\) denote the shared history of the first \(k\) evaluations, where \(y_i\) is the possibly noisy observed value and \(s_i\) records the evaluation status, such as success or failure. After \(t\) agent–environment interactions, let \(m_t\) denote the total number of objective evaluations, so that the corresponding history is \(H_{m_t}\).
Let $S_t$ contain the remaining budget and the numerical state of each method, and let $C_t$ denote the conversation available to the agent, including the task instructions and previous tool responses. With LLM $\mathcal{M}$ and harness $h$, the agent's
policy $\pi_{h,\mathcal{M}}$ interacts with the environment $E$ as follows:
\begin{equation}
 a_t\sim\pi_{h,\mathcal{M}}(\cdot\mid C_t),\qquad
 (o_{t+1},H_{m_{t+1}},S_{t+1})=E(H_{m_t},S_t,a_t).
 \label{eq:interaction}
\end{equation}
Here, $a_t$ is the tool request and $o_{t+1}$ is the returned feedback.
Specifically, once the agent selects a search tool and its parameters,
the tool computes candidate points using the evaluation history and its
retained numerical state. Let $j_t$ and $\lambda_t$ denote the tool and
parameters specified by $a_t$. The next point to be evaluated is given by
\begin{equation}
 x_{k+1}=\mathcal{N}_{j_t}(H_k,Z_k;\lambda_t),\qquad
 m_t\le k<m_{t+1},
 \label{eq:candidate-point}
\end{equation}
where $k$ counts objective evaluations, $\mathcal{N}_{j_t}$ denotes the
numerical method used by the selected tool, and $Z_k$ is its numerical state before
evaluation $k+1$.
During a multi-step call, the tool updates its state and generates further
points without requiring a new agent decision after every evaluation.
For an analysis call, $m_{t+1}=m_t$ and no new objective evaluation is made. Next, we describe the tools through which the agent analyzes the available
data and carries out numerical search.

\subsection{Analysis and Search Tools}
\label{sec:tools}
We provide two groups of tools: \emph{analysis tools} for examining existing
observations and local models, and \emph{search tools} for generating and evaluating candidate points.
These tools allow the agent to use the available evidence to select numerical methods and their parameters, and to revise these choices as new observations become available.

\paragraph{Analysis tools.}
Analysis tools assess search progress and the reliability of local models to
inform the agent's next decision. Generally, the best observed value alone is insufficient for deciding whether to continue a method, change its parameters, or improve its local model. To support these decisions,
\code{get\_situation} summarizes recent progress, remaining budget, search
scales, and local model readiness. For a closer examination, \code{preview\_model}
fits a local model and inspects its candidate without evaluating it. These
analyses use no objective evaluations and explicitly mark missing or unreliable
information. For example, evidence that a local model is poorly supported can lead
the agent to request additional samples before relying on a model-based step.

\paragraph{Search tools.}
Search tools execute numerical methods to generate and evaluate candidate points to improve the
objective or obtain samples needed for subsequent model-based search.
For example, the trust-region method seeks improvement using a local
quadratic model fitted by
minimum-Frobenius-norm-change interpolation
\citep{powell2004frobenius,powell2006newuoa}. It computes a trial step and
compares the actual and predicted reductions after evaluating the candidate.
When a useful local model is unavailable, poll search offers an alternative
by evaluating points along search directions. Geometry repair instead obtains
additional samples to improve the local interpolation set, preparing it for
later model-based steps.

The agent must also decide when to review a numerical method's progress.
Frequent feedback allows earlier intervention, while longer runs let the
method apply its own updates without repeated agent decisions. We therefore
provide two ways to invoke search tools. A \emph{single-step call}
performs one numerical operation, which may itself require several objective
evaluations. A \emph{multi-step call} runs a method for up
to an agent-specified number of objective evaluations before returning control,
with this allowance capped by the remaining global budget.
After either type of call, feedback reports what changed, how many evaluations
were used, and why the operation stopped or was rejected. The agent can then
decide whether to continue, reconfigure, or switch methods based on the observed
progress. To help the agent formulate valid requests, tool instructions specify
parameters, prerequisites, and evaluation cost;
Appendix~\ref{app:tool-catalog} (Table~\ref{tab:tools}) lists the available tools.

\subsection{State Management and Continuation}
State management lets a numerical method resume after a tool switch while
reusing earlier numerical work. Resuming a method requires suitable parameters,
such as its search scale, and a valid numerical state. In a conventional solver
built on a fixed algorithm, the parameters and numerical state evolve
according to that algorithm's update rules.
However, the NextEval solver needs to additionally determine
what to retain, adjust, or initialize when moving between methods. This makes
parameter selection and state management at tool switches an important part
of the solver design: the choices affect both where subsequent evaluations
are placed and how earlier numerical work can be reused.

To address this issue, we maintain a shared evaluation history while storing
each numerical method's state separately. The agent can use the history and
tool feedback to choose parameters, while the environment retains each method's parameters, including trust-region radii, polling scales, and step sizes for each coordinate, along with its numerical state. In particular, these parameters have different meanings and update rules, so they are not treated as interchangeable.
Moreover, retaining the numerical state does not mean that it can always be reused
unchanged. For example, if another method finds a better point, the saved
trust-region interpolation set may need to be updated. Accordingly, the tool  checks its set against the incumbent and can update or rebuild it when needed. Thus, continuation preserves earlier work while accounting for changes made by other tools.
Finally, the environment stores evaluation history and numerical state separately from the agent’s conversation, allowing numerical methods to continue even when the conversation is compressed.

\section{NextEval Bench: From One Function to a Suite}
\label{sec:bench}
\begin{figure}[htbp]
\centering
\includegraphics[width=\linewidth]{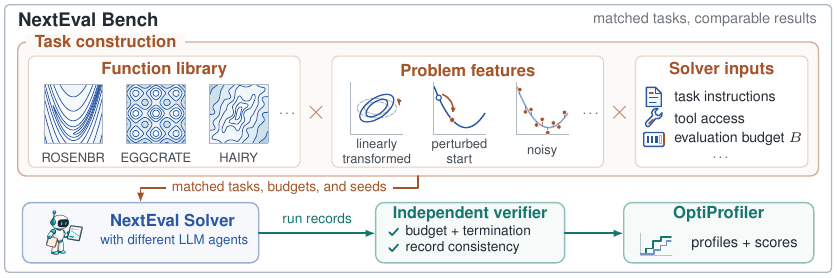}
\caption{NextEval Bench runs agent configurations on matched tasks through
the shared numerical environment and verifies run records before comparing
performance using OptiProfiler.}
\label{fig:bench}
\end{figure}

In this section, we introduce NextEval Bench (Figure~\ref{fig:bench}), which provides a protocol for evaluating how effectively different
agents use provided tools to solve the DFO problem~(\ref{eq:problem}). Building on the numerical environment introduced in Section~\ref{sec:solver}, the central idea is to vary the LLM or harness while keeping tasks, numerical tools, evaluation budgets, and permitted information fixed within each comparison. To do this, we specify how tasks are constructed, how agents interact with the environment, and how their optimization performance is evaluated.

\paragraph{Tasks and problem features.}
To evaluate agent performance across diverse DFO tasks, we use standard optimization test problems available through OptiProfiler \citep{optiprofiler}. Moreover, we extend the evaluation beyond their standard settings by perturbing the starting points, adding observation noise, and rotating the coordinates. For an objective function $f:\mathbb R^n\to\mathbb R$ and initial point $x_0\in\mathbb R^n$, these problem features are defined as follows:
\begin{align}
\text{Perturbed start:}\quad
 \widetilde{x}_0 &= x_0+\ell\max\{1,\|x_0\|\}\frac{u}{\|u\|},
 &u&\sim\mathcal N(0,I_n),\label{eq:perturbed-start}\\
\text{Noisy evaluation:}\quad
 y_i &= f(x_i)+\sigma\max\{1,|f(x_i)|\}\xi_i,
 &\xi_i&\overset{\mathrm{iid}}{\sim}\mathcal N(0,1),\label{eq:noisy-evaluation}\\
\text{Rotated coordinates:}\quad
 g(z)&=f(Qz),\qquad z_0=Q^\top\widetilde{x}_0,
 &Q^\top Q&=I_n.\label{eq:rotation}
\end{align}
Here, $\ell,\sigma\geq0$ are the start perturbation and noise levels, and $I_n$ is the identity matrix.
The Gaussian vector $u$ sets the perturbation direction; $x_i$ and $y_i$
are the point and observation at evaluation $i$, with fresh noise $\xi_i$
on every call, including repeated evaluations at the same point.
For rotation, the orthogonal matrix $Q\in\mathbb R^{n\times n}$ maps
$z\in\mathbb R^n$ to the original coordinates by $x=Qz$, so $g$ is the
transformed objective and $z_0$ its initial point.
We sample $u$ and $Q$ once per problem and repetition, obtaining $Q$ by orthogonalizing a Gaussian matrix, and share feature seeds across agents.

\paragraph{Agent configurations and execution.}
We evaluate a complete optimization session because each decision can depend
on earlier observations. A submission specifies the LLM, harness, prompts,
inference settings, and resource limits. Within each session, the harness
runs the interaction loop in Section~\ref{sec:solver}, while the numerical environment preserves
evaluation history and numerical state. Objective evaluations, tool calls,
LLM calls, and wall time are recorded separately. Problem identities,
formulas, and scoring references are not exposed to the agent.

\paragraph{Performance evaluation.}
Performance is measured by the evaluations needed to reach relative reduction
targets, so results can be compared across problems with different objective
scales. We verify run settings, evaluation counts, termination, and record
consistency before constructing OptiProfiler performance and data profiles;
failed and retried attempts remain in the run records. Where retries occur,
profiles use the final authorized attempt, not the best attempt, and do not
sum evaluation costs across attempts.
For noisy tasks, clean objective values are reserved for scoring.
Section~\ref{sec:experimental-setup} and Appendix~\ref{app:score} give the
scoring definitions.

The agent-configuration studies in Section~\ref{sec:configuration-studies} follow these comparison principles to evaluate optimization performance under matched tasks, numerical tools, and evaluation budgets.
\vspace{-3pt}
\section{Experiments}
\label{sec:experiments}
In this section, we first compare the NextEval solver with classical solvers. We then audit one recorded trajectory to illustrate how the LLM
agent adapts its search. Finally, we study the influence of agent configuration through controlled comparisons following the principles of NextEval Bench. Table~\ref{tab:studies} summarizes the questions we aim to answer through these experiments.
\vspace{-6pt}
\begin{table}[ht]
\centering\small
\setlength{\tabcolsep}{6pt}
\renewcommand{\arraystretch}{1.02}
\caption{Research questions guiding the experiments.}
\label{tab:studies}
\begin{tabularx}{\linewidth}{@{}
>{\raggedright\arraybackslash}p{0.22\linewidth}
>{\raggedright\arraybackslash}X@{}}
\toprule
\textbf{Study} & \textbf{Research question}\\
\midrule
Solver performance & Can the NextEval solver improve evaluation efficiency over classical DFO solvers?\\
\addlinespace[1pt]
Case study & How does the agent adapt its search in one audited trajectory?\\
\addlinespace[1pt]
Model choice & How does the choice of LLMs affect optimization performance?\\
\addlinespace[1pt]
Reasoning setting & Does additional reasoning improve optimization performance?\\
\addlinespace[1pt]
Execution harness & How sensitive is optimization performance to the execution harness?\\
\bottomrule
\end{tabularx}
\end{table}

\subsection{Experimental Setup}
\label{sec:experimental-setup}
We use a common suite comprising all 92 unconstrained problems with dimension
$n\leq5$ in the archived S2MPJ collection \citep{gratton2025s2mpj}.
The suite spans classical nonlinear optimization, nonlinear least-squares, and data-fitting problems.
All runs use a budget of $50n$ objective evaluations.
Using the features defined in
Eqs.~(\ref{eq:perturbed-start})--(\ref{eq:rotation}), we test perturbed
starts alone, with noise, and with rotation but no noise, setting
$\ell=\sigma=10^{-3}$.
Appendix~\ref{app:protocol}
gives protocol details and agent information restrictions.

\paragraph{Performance and data profiles.}
\label{sec:metrics}
Because objective scales differ across problems, we compare evaluations
needed to achieve a relative reduction in objective value.
For each problem and repetition, we define a \emph{target}, an objective-value
threshold. At tolerance $\tau$, the remaining gap to the best noise-free
value found by the compared solvers must be at most $\tau$ times the initial gap.
We summarize the first evaluation counts reaching these targets using
performance profiles $\rho_s$ \citep{dolan2002profiles} and data profiles $\delta_s$
\citep{morewild2009benchmarking}, as implemented in OptiProfiler
\citep{optiprofiler}.
Performance profiles normalize these counts by the smallest count on each
problem, whereas data profiles scale them by the dimension plus one;
higher curves indicate more targets reached within the respective budget.

At tolerance $\tau_k=10^{-k}$, let $\bar\rho_{s,k}$ and $\bar\delta_{s,k}$
denote profiles averaged over repetitions. On the plotted logarithmic axes,
$\widetilde\rho_{s,k}(x)=\bar\rho_{s,k}(2^x)$ and
$\widetilde\delta_{s,k}(x)=\bar\delta_{s,k}(2^x-1)$.
For either plotted profile $\phi_{s,k}$, its area and normalized score are
{\small
\setlength{\abovedisplayskip}{4pt}\setlength{\belowdisplayskip}{4pt}
\setlength{\abovedisplayshortskip}{4pt}\setlength{\belowdisplayshortskip}{4pt}
\[
\mathrm{AUC}(\phi)=\int_a^b\phi(x)\,\mathrm{d}x,
\qquad S_{\mathrm{OP}}(\phi_{s,k})=
\frac{\mathrm{AUC}(\phi_{s,k})}{\max_j\mathrm{AUC}(\phi_{j,k})}.
\]
}
The integration interval $[a,b]$ is shared across solvers for each profile type
and tolerance. The aggregate score averages the performance-profile scores
over the ten tolerances:
$S_s=\frac{1}{10}\sum_{k=1}^{10}S_{\mathrm{OP}}(\widetilde\rho_{s,k})$.
The figures show profiles at tolerance $\tau=10^{-6}$; Table~\ref{tab:scores}
also reports scores at $\tau=10^{-4}$.
Because the reference value and score normalization depend on the compared
solvers, scores from different solver sets are not directly comparable.
Noisy studies are scored using noise-free objective values.
Appendix~\ref{app:score} gives the exact definitions and scoring conventions.

\raggedbottom
\subsection{Primary Performance Comparison with Classical DFO}
\label{sec:r2}

The primary rotated comparison uses NextEval's built-in runner and DeepSeek V4 Flash
(thinking disabled) with ten repetitions. The earlier
perturbed-start cohorts, with and without noise, use the historical task-mode
runner with three repetitions and are shown only as a
feature-suite overview; they are not pooled with the rotated comparison.

\begin{samepage}
The reference solvers are standalone COBYQA 1.1.4
\citep{ragonneau2022thesis,ragonneau2026cobyqa}, NEWUOA through PDFO 2.2.0
\citep{powell2006newuoa,ragonneau2024pdfo}, SciPy's implementation of the modified Powell algorithm,
and SciPy BFGS with forward finite differences \citep{virtanen2020scipy}.
All solver evaluations, including finite-difference and line-search probes,
count toward the same $50n$ objective budget.
Implementation details are given in Appendix~\ref{app:protocol}.
\par
\end{samepage}

\begin{figure}[t]
\centering
\setlength{\tabcolsep}{1.5pt}
\begin{tabular}{@{}ccc@{}}
\small Perturbed starts & \small Perturbed starts + noise & \small Perturbed starts + rotation\\[2pt]
\includegraphics[width=.32\linewidth]{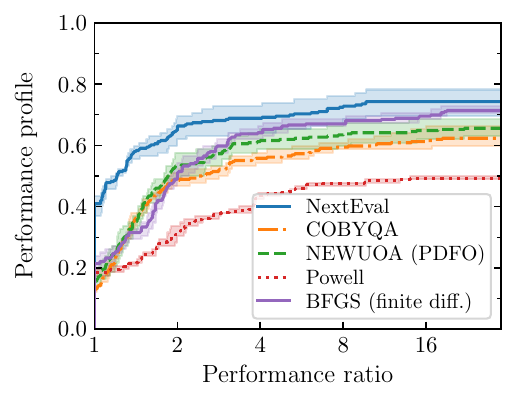} &
\includegraphics[width=.32\linewidth]{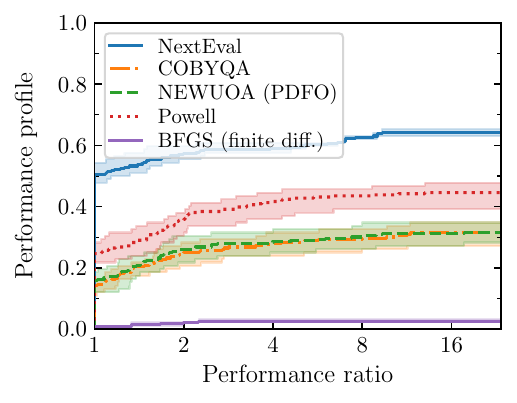} &
\includegraphics[width=.32\linewidth]{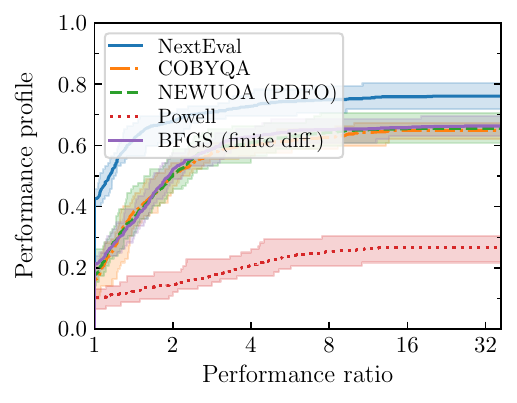}\\[-1pt]
\includegraphics[width=.32\linewidth]{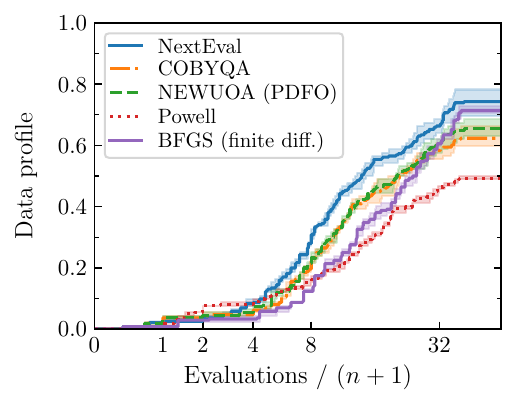} &
\includegraphics[width=.32\linewidth]{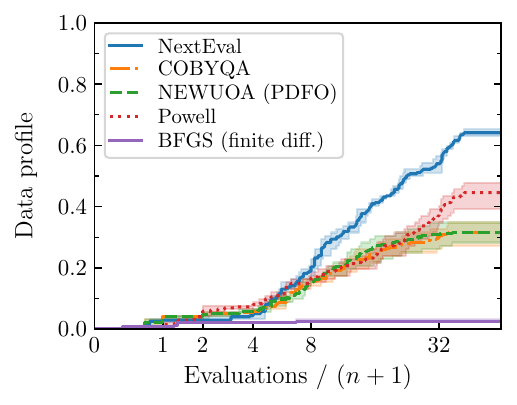} &
\includegraphics[width=.32\linewidth]{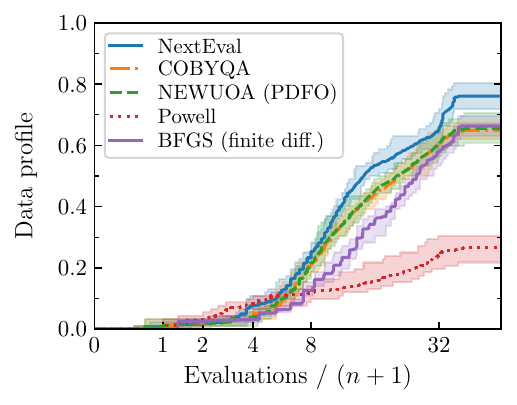}
\end{tabular}
\caption{\textbf{DeepSeek V4 Flash feature-suite overview.}
History-performance (top) and history-data (bottom) profiles at $10^{-6}$;
shading shows the repetition range.
The first two columns use historical three-repeat cohorts; rotation uses the
built-in runner with thinking off and ten repeats. These are not matched
feature ablations.}

\label{fig:profiles}
\end{figure}

NextEval has the largest aggregate score in the rotated comparison
(Table~\ref{tab:scores}; Figure~\ref{fig:profiles}). This measures the complete system: agent
decisions together with numerical tools. Finite-difference BFGS is
competitive without noise but degrades under noise with its default
difference step. Appendix~\ref{app:completion} gives trial accounting.

\Needspace{21\baselineskip}
\subsection{Audited Case Study}
\label{sec:case-study}
\begin{wrapfigure}[18]{r}{.46\textwidth}
\vspace{-7pt}
\centering
\includegraphics[width=.9\linewidth]{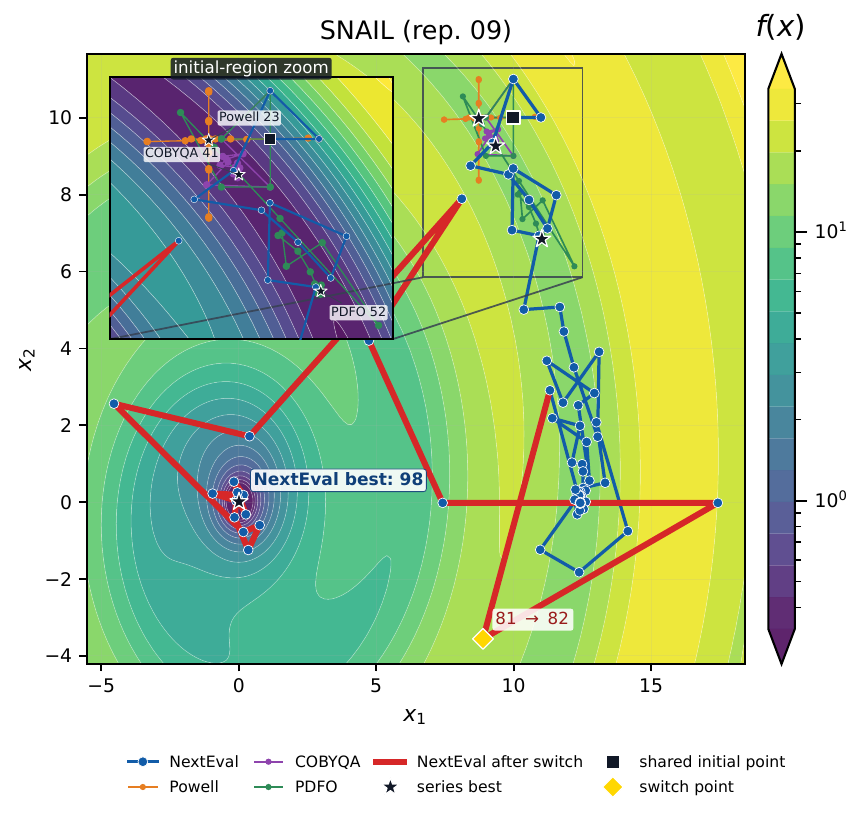}
\small
\setlength{\abovecaptionskip}{4pt}
\caption{Recorded \textsc{SNAIL} trajectories (repetition 9). Red: evaluations 81--98. Inset: shared initial region. Clean contours are evaluator-only.}
\label{fig:snail-case-main}
\vspace{-7pt}
\end{wrapfigure}
To make the aggregate profiles concrete, we audit one recorded trajectory. In the model-choice study of Section~\ref{sec:configuration-studies} (perturbed starts plus noise, ten repetitions), GPT-5.6 Terra under the Codex harness stalled on \textsc{SNAIL} (repetition 9), with the incumbent remaining at $13.38$ through evaluations 78--81.
 After evaluation 81, the agent switched to a reset trust-region campaign with radius factor 5.0, reaching a best recorded value of \(1.157\times10^{-3}\) at evaluation 98 within the 100-evaluation budget. The corresponding clean objective value was $2.101\times10^{-4}$.
The clean values and contours were checks performed by the evaluator and were unavailable to the agent, allowing the trajectory to be inspected without exposing this information during optimization. Appendix~\ref{app:cases} provides the full trace reconstruction, screening details, and additional trajectories.
\WFclear

\begin{table}[t]
\centering\small
\setlength{\tabcolsep}{5pt}
\renewcommand{\arraystretch}{1.02}
\setlength{\aboverulesep}{2pt}\setlength{\belowrulesep}{2pt}
\caption{Normalized profile-area scores for perturbed starts with rotation.
Perf. and Data report $S_{\mathrm{OP}}(\widetilde\rho_{s,k})$ and
$S_{\mathrm{OP}}(\widetilde\delta_{s,k})$; Aggregate score reports $S_s$.
Higher is better; scores are not win rates.}
\label{tab:scores}
\begin{tabular*}{\linewidth}{@{\extracolsep{\fill}}lccccc@{}}
\toprule
Solver & Aggregate score & Perf. $10^{-4}$ & Data $10^{-4}$ & Perf. $10^{-6}$ & Data $10^{-6}$\\
\midrule
NextEval & 1.000 & 1.000 & 1.000 & 1.000 & 1.000\\
COBYQA & 0.805 & 0.847 & 0.899 & 0.796 & 0.856\\
NEWUOA (PDFO) & 0.820 & 0.862 & 0.913 & 0.807 & 0.864\\
Powell (SciPy) & 0.354 & 0.370 & 0.466 & 0.300 & 0.393\\
BFGS (finite differences) & 0.810 & 0.843 & 0.793 & 0.818 & 0.756\\
\bottomrule
\end{tabular*}
\end{table}

\subsection{Controlled Agent-Configuration Studies}
\label{sec:configuration-studies}
\begin{figure}[htb]
\centering
\setlength{\tabcolsep}{1.5pt}
\begin{tabular}{@{}ccc@{}}
\small (a) GPT: model choice & \small (b) Qwen: thinking & \small (c) DeepSeek: harness\\[2pt]
\includegraphics[width=.325\linewidth]{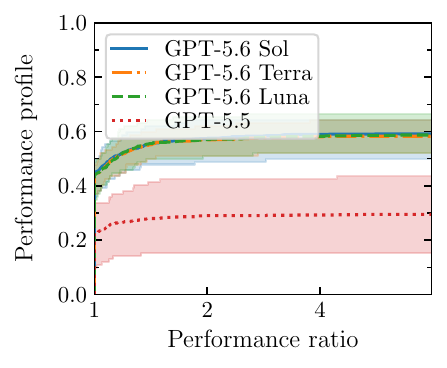} &
\includegraphics[width=.325\linewidth]{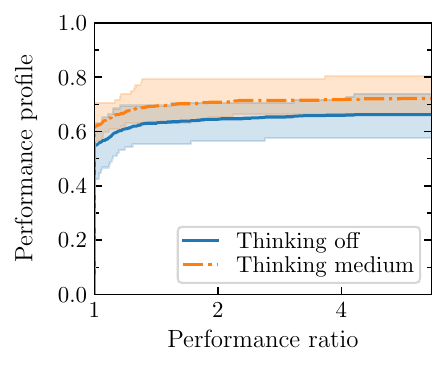} &
\includegraphics[width=.325\linewidth]{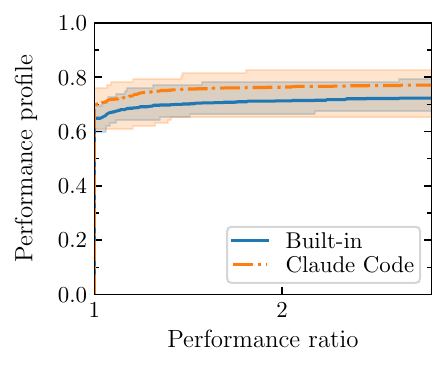}
\end{tabular}
\caption{\textbf{Three agent-configuration comparisons.} History-performance
profiles at $10^{-6}$, with repetition ranges. (a) Codex with medium reasoning;
(b) Qwen 3.8 Flash with Codex; (c) DeepSeek V4 Flash with thinking off.
All three comparisons use the full 92-problem suite.
Targets are computed separately within each panel. Data profiles
and trial accounting are in Appendices~\ref{app:ten-repeat}--\ref{app:harness-study}.}
\label{fig:agent-studies}
\end{figure}

We compare model choice, reasoning setting, and execution harness under
perturbed starts plus noise, with ten repetitions in each study
(Figure~\ref{fig:agent-studies}). These comparisons use matched tasks
within each study, not a pooled ranking across the three studies.

\paragraph{Model choice (GPT).}
\label{sec:gpt}
With Codex and medium reasoning fixed, GPT-5.6 Sol, Terra, and Luna form a
tight group at $10^{-6}$ and reach more targets than GPT-5.5. Their aggregate
scores range from 0.981 to 0.997, versus 0.571 for GPT-5.5. However, GPT-5.5
uses only 13\% of the budget at the median, versus 100\% for GPT-5.6.
The difference therefore reflects both search and stopping behavior under
the same harness, rather than isolating reasoning ability.

\paragraph{Reasoning setting (Qwen).}
\label{sec:thinking}
With Qwen 3.8 Flash and Codex fixed, medium thinking improves the aggregate
score from 0.924 to 1.000 and obtains a lower median final clean value on
53 problems, versus 39 for thinking off. Additional reasoning helps in
this setting, although it does not improve every problem.

\paragraph{Execution harness (DeepSeek).}
With DeepSeek V4 Flash, thinking off, and the same NextEval instructions,
18 tools, decoding limits, and evaluation budgets, Claude Code scores
0.999 versus 0.947 for the built-in runner. Partial failure histories are
retained. This measures the effect of the complete harness, including
context management and session recovery, not one internal component.

\section{Related Work}
\label{sec:related}
\paragraph{Numerical optimization without derivatives.}
Direct search, interpolation-based trust regions, evolution strategies,
and Bayesian optimization (BO) provide established approaches to DFO
\citep{kolda2003,conn2009idfo,hansen2016tutorial,shahriari2016loop}.
NextEval exposes numerical operations as tools rather than replacing their
calculations with language-model arithmetic. TuRBO also studies adaptive
allocation \citep{rw20260906_eriksson_turbo}, but its allocation rule is part
of a specified optimization algorithm.

\paragraph{Language models and learned optimizers.}
Learned recurrent DFO already adapts to evaluation histories
\citep{rw20260906_chen_learned_dfo}.
OPRO generates candidates from past evaluations, and EvoLLM uses LLMs for
population recombination
\citep{rw20260906_yang_opro,rw20260906_lange_evollm}.
LLAMBO implements initialization, surrogate prediction, and candidate
sampling within BO without fine-tuning \citep{rw20260906_liu_llambo}.
SemanticOpt trains on BO trajectories with semantic context to generate
configurations and predictions \citep{rw20260912_meindl_semanticopt}.
These algorithms use feedback from earlier evaluations, but their learned or
prompted component primarily generates candidates, configurations, or
surrogate quantities. NextEval instead gives the agent an operational choice:
inspect the history, select an implemented method, set its parameters,
continue or switch it, and allocate bounded evaluations. The distinction is
the controlled action space, not merely the presence of a language model.

\paragraph{Optimization agents.}
Code-to-Harness evolves executable optimizers, distills their practice into
a short text strategy, and freezes it before deployment
\citep{wu2026codetoharness}. Its executor uses evaluation history, including
in sandbox-equipped experiments; it is not a solution-mapping approach.
It discovers portable strategies, whereas NextEval provides stateful
numerical tools for online control and a protocol for comparing agents.
A distilled strategy could guide an agent using NextEval tools.
AgentHPO coordinates hyperparameter proposals, experiment execution, and
log analysis \citep{rw20260906_liu_agenthpo}. Multi-agent BO separates an
agent that weights search criteria from an agent that generates candidates
\citep{rw20260906_carbonati_multiagent}; OptAgent adds a related agentic
layer to physics-grounded building simulations \citep{rw20260906_jiang_optagent_buildings}.
Related work from Bin Dong's group studies reusable tool graphs in GATE
\citep{rw20260921_luo_gate} and multi-round LLM interaction through optimal
control \citep{rw20260921_luo_promptcontrol}. Iteris applies an
explore--plan--execute loop to computational mathematics
\citep{rw20260921_chen_iteris}. These works motivate tool-based control,
but do not study the same oracle-only DFO setting and evaluation budget.
Appendix~\ref{app:design-rationale} discusses this design choice and adjacent
approaches.

\paragraph{Evaluation infrastructure.}
COCO and OptiProfiler standardize numerical comparisons
\citep{hansen2021coco,optiprofiler}; MetaBox-v2 studies learning-based
DFO algorithms \citep{rw20260906_ma_metaboxv2}.
Harbor specifies task and agent trajectory formats
\citep{harbor2026atif,harbor2026tasks}, while Terminal-Bench-Science provides
scientific research tasks for evaluating agents \citep{tbscience2026}.
NextEval Bench combines oracle-controlled numerical tasks, agent-level
records, and suite-level profiles. Task and participant manifests fix the
evaluation conditions and complete LLM--harness configuration; the
current scope is a protocol and local execution machinery, not hosted
benchmarking.

\section{Limitations and Benchmark Roadmap}
Our primary comparison uses public problems with at most five variables.
Development feedback informed the tools, manuals, and prompts, so
generalization to unseen families remains untested. Perturbed starts,
rotations, and noisy observations do not rule out problem recognition.
The rotation used here preserves the objective's value structure and is not
full anonymization. We measure evaluation
counts, not total runtime or LLM cost. Whole-system comparisons do not
attribute gains to the LLM alone. Appendix~\ref{app:ablation-design}
separates controller replacement from numerical-support ablations; these
are planned tests, not completed results. Training a parameterized controller,
including by reinforcement learning, is a separate extension. Higher-dimensional
and additional problem families remain important tests.

\section{Conclusion}
NextEval places a persistent language-model agent in a conventional DFO solver;
the agent selects history summaries and numerical methods while the environment
executes evaluations and retains method state. On the low-dimensional S2MPJ
suite with perturbed starts and rotated coordinates, the primary configuration
has a higher aggregate score than the tested DFO baselines.
NextEval Bench compares LLM--harness configurations under
a common evaluation budget.
\label{main-text-end}

\clearpage
\flushbottom
\subsection*{Reproducibility Statement}
All displayed profiles are generated by OptiProfiler from saved
histories. Appendix~\ref{app:protocol} specifies experimental settings;
Appendix~\ref{app:completion} records trial accounting. Historical
experiments are reported under the settings used when they were conducted.
\subsection*{AI Use Statement}
AI assistants supported coding, literature discovery, experimental-design
feedback, qualitative trajectory analysis and interpretation, drafting,
language editing, and creation of Figures~\ref{fig:concept}--\ref{fig:bench}.
The authors are responsible for verifying the final code, citations, and empirical claims.
\subsubsection*{Acknowledgments}
The work of Li Cui was supported by the National Natural Science Foundation of
China under Grant 12171043.

The work of Cunxin Huang and Zaikun Zhang was supported by the Guangdong Major
Project of Basic Research (2026B0303000015), NSFC General Program (12571335),
and NSFC Excellent Young Scientists Fund (Overseas). Cunxin Huang was also
supported by the Hong Kong PhD Fellowship Scheme (PF21-56718).

\clearpage
\appendix
\section{Experimental Protocol}
\label{app:protocol}
The studies use OptiProfiler 1.3.4 and the common suite defined in
Section~\ref{sec:experimental-setup}. Historical studies are described
separately below.
Representative instances include ROSENBR,
HAIRY, EGGCRATE, BARD, MEYER3, and CHWIRUT1LS, spanning classical nonlinear
optimization, nonlinear least-squares, and data-fitting problems. All solver
runs are capped at $50n$ objective calls, implemented as
\code{max\_eval\_factor}=50 in OptiProfiler. Each agent occupies one persistent
session per problem and repetition. In task mode, one runner owns the entire
solve. Model calls, tool calls, and
objective calls are distinct quantities. The primary study is the rotated
solver comparison, while the main DeepSeek V4 Flash feature suite also
includes the historical perturbed-start profiles, with and without noise. The remaining studies compare
agent configurations under their respective experimental settings. These cohorts are not
pooled. Saved HDF5 histories are reloaded to generate all profiles.
Completed trials, including normal early stops and verified session-limit
stops, are retained. Failed attempts may be retried under the recorded
recovery policy. The selected result is the latest authorized terminal
attempt, not the attempt with the best objective value; completed parents
are not retried. All attempts remain in the audit trail.

\begin{table}[ht]
\centering\small
\caption{Execution settings. Wall limits are
operational caps; none of the paper's scores use elapsed time.}
\label{tab:execution-settings}
\begin{tabular}{@{}>{\raggedright\arraybackslash}p{2.7cm}
>{\raggedright\arraybackslash}p{4.8cm}
>{\raggedright\arraybackslash}p{5.0cm}@{}}
\toprule
Field & DeepSeek Flash feature suite & Model configuration studies\\
\midrule
Repetitions & Earlier cohorts: 3; rotated: 10 & Ten in each study\\
Features & Perturbed starts alone, with noise, or with rotation; start/noise levels $10^{-3}$ & Perturbed starts with noise; both levels $10^{-3}$\\
Harness & Earlier cohorts: historical task-mode; rotated: built-in task-mode runner & GPT and Qwen thinking: Codex; harness study: built-in vs. Claude Code\\
Models & DeepSeek V4 Flash (observed ID below) & GPT configurations, Qwen 3.8 Flash, and DeepSeek V4 Flash\\
Thinking & Earlier cohorts: historical setting; rotated: disabled & GPT: medium; Qwen thinking: off/medium; harness comparison: off\\
Session wall limit & Earlier cohorts: 1,800 seconds; rotated: 3,600 seconds & GPT: 1,800 seconds; Qwen thinking: no solver-side wall cap; harness: 3,600 seconds\\
Other limits & Earlier cohorts: historical; rotated: 200 turns and 200 tool calls & Harness: 200 turns and 200 tool calls; other studies: recorded runner options\\
Fallback solver & Earlier cohorts: historical; rotated: none & None in all three studies\\
Context & Earlier cohorts: historical digest; rotated: persistent session and debug trace & Harness-specific, held fixed within each model study\\
Reference solvers & COBYQA, NEWUOA (PDFO), Powell (SciPy), finite-difference BFGS (SciPy); versions below & Agent-only profiles; archived classical runs used PDFO automatic dispatch\\
\bottomrule
\end{tabular}
\end{table}

\paragraph{Feature definitions.}
Equations~\eqref{eq:perturbed-start}--\eqref{eq:rotation} define the three
features. The start direction and, when applicable, orthogonal matrix are
sampled once per problem--repetition pair and shared across solvers. Random
noise is sampled on each evaluation, including repeated calls at the same
point. Clean values are reserved for scoring. Only start perturbations and
noise have level $10^{-3}$; rotation is an orthogonal change of coordinates,
without anisotropic scaling or observation noise.

The earlier perturbed-start cohorts, with and without noise, use the
archived task-mode runner with three repetitions.
The primary rotated cohort uses the built-in task-mode runner with
DeepSeek V4 Flash (thinking disabled) and ten repetitions.
Because these cohorts differ in runner and numerical implementations, as well
as repetition scheme, they are analyzed separately and are not pooled. The rotated suite uses the
observed model path \path{deepseek-v4-flash-ga-260731}. Its feature
seeds are $0,211,422,\ldots,1899$; each problem--repetition pair has a
matched perturbed start and orthogonal rotation, and no observation noise.
All five displayed solver arms receive the same transformed start and $50n$ objective
budget. The ten-repeat GPT study uses feature seeds
$0,211,422,\ldots,1899$ and solver seeds $0,\ldots,9$.
The historical three-repeat GPT campaign uses only
the first three seed pairs.

The historical seven-arm GPT and Qwen campaigns called PDFO 2.2.0 without a method argument.
In the historical GPT campaign, its automatic dispatch used UOBYQA on the 90 problems
with dimensions two through five and NEWUOA on the single one-dimensional
problem. The DeepSeek Flash solver comparisons explicitly requested NEWUOA
through PDFO. These labels
identify the software that generated the reported histories, not merely the
underlying Powell methods; no PRIMA implementation was used in these studies.

\paragraph{Supplementary classical baselines.}
The displayed reference set uses standalone COBYQA 1.1.4, NEWUOA through PDFO 2.2.0,
SciPy's modified Powell direction-set method, and BFGS with forward finite
differences. Powell here names SciPy's implementation, not the family that
also includes NEWUOA. The primary rotated comparison uses SciPy 1.15.3 for
both Powell and the added BFGS arm. The earlier two cohorts were supplemented
with Powell and BFGS using SciPy 1.18.1, without new agent calls.
The supplements match the archived problems, dimensions, feature seeds,
transformed starts, and $50n$ budgets; the retained solver histories are
unchanged.
BFGS uses \code{jac="2-point"}, the default automatic relative difference
step, and otherwise default SciPy settings. Powell uses default settings
with \code{maxfev}=50n. An objective wrapper stops before any call beyond
the budget, including finite-difference probes and line-search evaluations.
At a forced BFGS budget stop, it returns the best point according to its
observed values. Clean values are available only to the evaluator.
This standard BFGS baseline is not tuned for noisy observations: a small
difference step can amplify noise, so its noisy performance should not be
interpreted as that of a noise-adapted gradient estimator.
SciPy Powell returns a nonfinite final point in six supplementary trials
without noise and fourteen with noise. These runs are not discarded or
restarted; their recorded evaluations remain in the history-based profiles.

The original GPT campaign runs inside WSL2 Linux, not Windows-hosted
Python. Its Codex configuration uses medium effort and sets
\code{code\_mode\_host=True}; the recorded runner disables external host
shell, execution, and code tools.
The primary DeepSeek Flash runner disables built-in tools and exposes only
the declared NextEval tools through MCP. The 40 MISRA1ALS GPT supplement
episodes use the same numerical implementation and runner adapter on
macOS with Codex CLI 0.154.0; the original Windows episodes used CLI 0.153.4.
All four supplement arms share the new host and version. This platform
deviation is recorded rather than treated as exact same-host replication.
The environment records tool and
objective calls and enforces the $50n$ objective-evaluation budget. The Qwen Codex runs disable shell and web-search tools,
use a read-only sandbox, and expose NextEval tools through MCP. The Qwen
thinking study has no solver-side wall-clock cap; it instead limits each
episode to 60 provider requests, two same-thread resumptions, and 200 task
tool calls. These are distinct from its $50n$ objective-evaluation budget.
None of the primary, GPT, or Qwen studies passes unused budget to a fallback
solver. In the primary study, three nonfinite MISRA1ALS starting points
were replaced using prespecified reserve seeds; only the start perturbation
seed changed, while the rotation seed was preserved. The historical GPT
archive excluded MISRA1ALS; its 92-problem four-model comparison adds ten
new episodes per model, using the same finite-start substitutions while
retaining the original noise seeds. The primary and GPT dimension histograms are
1/41/27/11/12 for dimensions 1/2/3/4/5.

\subsection{Agent Information and Problem Recognition}
\label{app:information}
The numerical environment retains the objective callable. Its task payload
provides the runner with the dimension, initial point and observed value,
budgets, and tool descriptions, but no benchmark identifier, function
formula, reference optimum, or evaluator output path.
The historical task-mode prompt additionally instructs the
agent to base decisions on tool observations rather than memories of named
test functions. This is an instruction, not an enforceable restriction on
the model's internal reasoning.
The perturbations and observation noise reduce reliance on familiar starting
points and exact function values, but they are not intended as a guarantee
against problem recognition.
The observed points and values may still reveal the function family.
In particular, changing the starting point leaves the objective unchanged,
and the noise used here has zero mean, so the minimizer of the clean
objective remains unchanged. These measures therefore do not certify
absence of problem recognition or training-data contamination.
The payload boundary also does not by itself establish filesystem
isolation for an external coding harness; the runner configurations and
tool restrictions are recorded above.

\subsection{Exact Profile Score and Nonfinite Conventions}
\label{app:score}
For a problem and repetition, let $f^{\rm ref}$ be the best clean value
found in the declared comparison set, including the initial value.
When both values are finite, the target at tolerance $\tau$ is
\begin{equation}
 f^{\rm tar}(\tau)=
 \max\{f^{\rm ref},\ \tau f(x_0)+(1-\tau)f^{\rm ref}\}.
 \label{eq:target}
\end{equation}
Following OptiProfiler \citep{optiprofiler}, let $t_{p,s}^{(r)}(\tau)$ be
the first evaluation at which solver $s$ reaches this target on problem $p$
in repetition $r$; it is infinite if the target is not reached. Suppressing
the tolerance and repetition indices, define the relative and dimension-scaled
costs by
\[
r_{p,s}=\frac{t_{p,s}}{\min_{j\in\mathcal S}t_{p,j}},
\qquad d_{p,s}=\frac{t_{p,s}}{n_p+1},
\]
where $\mathcal S$ is the compared solver set and $\infty/\infty=\infty$.
For problem set $\mathcal P$, the performance and data profiles are
\begin{equation}
\begin{aligned}
\rho_s(\alpha)&=\frac{|\{p\in\mathcal P:r_{p,s}\le\alpha\}|}{|\mathcal P|},
&&\alpha\ge1,\\
\delta_s(\alpha)&=\frac{|\{p\in\mathcal P:d_{p,s}\le\alpha\}|}{|\mathcal P|},
&&\alpha\ge0.
\end{aligned}
\end{equation}
Higher curves mean more targets reached within the relative or
dimension-scaled budget. At $\tau_k=10^{-k}$, averaging over $R$ repetitions gives
$\bar q_{s,k}(\alpha)=R^{-1}\sum_{r=1}^R q_{s,k}^{(r)}(\alpha)$
for either $q=\rho$ or $q=\delta$.

The plotted curves are $\widetilde\rho_{s,k}(x)=\bar\rho_{s,k}(2^x)$ and
$\widetilde\delta_{s,k}(x)=\bar\delta_{s,k}(2^x-1)$, corresponding to
$x=\log_2\alpha$ and $x=\log_2(1+\alpha)$, respectively.
For either plotted curve $\phi$, OptiProfiler integrates with unit weight over
$[a,b]=[0,1.1\max(x_{\max},\epsilon_{\rm mach})]$, where $x_{\max}$ is the
largest finite transformed attainment ratio in that profile comparison.
Unattained targets do not increase the curve. Including the interval endpoints
among the breakpoints $x_i$, the right-continuous step curve has exact area
\begin{equation}
\mathrm{AUC}(\phi)=\int_a^b\phi(x)\,\mathrm{d}x
=\sum_i(x_{i+1}-x_i)\phi(x_i).
\end{equation}
The normalized score $S_{\mathrm{OP}}(\phi_{s,k})$ divides this area by the
largest area among solvers at the same tolerance and profile type;
a zero denominator is replaced by one. Table~\ref{tab:scores}'s Perf. and Data
columns report $S_{\mathrm{OP}}(\widetilde\rho_{s,k})$ and
$S_{\mathrm{OP}}(\widetilde\delta_{s,k})$, respectively. Its Aggregate score is
$S_s=10^{-1}\sum_{k=1}^{10}S_{\mathrm{OP}}(\widetilde\rho_{s,k})$.
Thus repetition averaging precedes integration,
normalization occurs within each tolerance, and the ten normalized
scores are then averaged. The displayed $10^{-6}$ panels are not
selected separately for each model family.

The archived scorer maps NaN merit values to $+\infty$. With a finite
initial merit, an all-invalid trajectory cannot attain a finite target.
For an infinite initial merit, however, OptiProfiler's archived
convention uses an infinite target and declares all solvers to pass;
this degenerate case is not evidence of successful optimization.
The full-population plots preserve that convention. The finite-start
sensitivity analysis below removes the affected problem.

\section{Trial Accounting and Reference-Set Sensitivity}
\label{app:completion}

\subsection{Primary rotated suite}
The primary profiles retain the full problem suite, not only successful
agent sessions. The suite has 92 problems, ten repetitions,
five solver arms, and therefore 4,600 solver trials. Each arm uses the
same perturbed start followed by an orthogonal coordinate rotation, no
observation noise, and the same $50n$ objective budget. Every available
objective call remains charged to that budget. The final history-based
performance and data profiles are generated by reloading the recorded
evaluation histories in OptiProfiler.

Of the 920 agent trials, 914 completed and six failed: five on MGH17LS
(repetitions 0, 1, 2, 3, and 9). Four stopped after repeated invalid
evaluations and one after an eigensolver error. One MISRA1ALS trial
(repetition 7) stopped after repeated invalid evaluations. All six recorded histories
remain in the comparison; no failed agent trial was rerun or excluded.
The four classical arms each contribute 920 trials. The BFGS supplement
preserves all retained histories exactly and respects every $50n$ cap.
The runner used 1,024 output tokens per request, with truncation recovery.
The results characterize this runner configuration.

Table~\ref{tab:scores} reports the resulting five-arm comparison. The
feature replay audit verified the recorded rotated values against
$f(Qz)$ and checked the orthogonality of each stored transformation.
Because an orthogonal rotation preserves objective values and can preserve
recognizable structure, this audit supports feature identity, not a claim
that public test functions are unrecognizable to an agent.

\subsection{Earlier perturbed-start studies}
The following agent accounting belongs to the earlier development study,
not the primary rotated result. The refreshed profiles retain every
recorded problem and repetition, including interrupted or budget-limited
agent runs, and add the matched Powell and BFGS results. Recovered
transient errors do not automatically exclude an otherwise completed trial.

\begin{table}[ht]
\centering\small
\caption{Earlier perturbed-start study accounting after the classical-baseline supplement.
Agent events refer to the unchanged historical runs.}
\begin{tabular}{@{}lrr@{}}
\toprule
Quantity & Perturbed start & Perturbed start + noise\\
\midrule
Problems & 92 & 92\\
All displayed solver trials & 1,380 & 1,380\\
Agent trials & 276 & 276\\
Runner interruptions & 8 & 12\\
Session wall limits & 17 & 47\\
Invalid initial values & 2 & 2\\
Problems with nonfinite starts & 1 & 1\\
\bottomrule
\end{tabular}
\end{table}

The NextEval--DFO campaign resumes with 290 and 284 verified point-by-point
replays across all solvers, respectively. Replayed observed values are
checked against their originals. This completes the interrupted batch;
it does not add independent repetitions or produce valid fresh-run
timings. All 92 numerical histories are retained even where a historical
individual-plot export failed.

The previous four-arm completion-conditioned scores and bootstrap intervals
are not reused: changing the participant set changes the empirical targets
and profile normalization. The current profiles and aggregate scores are
recomputed for all five displayed solvers.

For the two earlier five-arm sets, we resample problems with replacement 5,000
times, retaining all solvers and repetitions of each sampled problem.
First-hit evaluation counts come directly from OptiProfiler; profile
normalization is recomputed for each draw. The 95\% percentile intervals
for NextEval's aggregate score minus the strongest reference in each draw
are $[0.010,0.217]$ for perturbed starts and $[0.178,0.477]$ with noise.
Excluding the single problem with nonfinite
perturbed starts from each earlier cohort preserves
NextEval's highest aggregate score. These intervals describe variation over
this problem suite, not performance on arbitrary objectives.

\subsection{Historical Three-Repeat GPT Snapshot}
\label{app:models}
This subsection retains the earlier three-repeat GPT analysis;
its scores and bootstrap intervals are not the ten-repeat main results.
The earlier GPT campaign contains 1,911 numerical cells across seven solvers;
the four-GPT view contains 1,092. An audit of the recorded trials
reports no unsuccessful final cells or output fallbacks. Nonfinite
evaluations remain counted. Replayed timing is not used for comparison.

The four-GPT history target counts at $10^{-6}$ are 171, 159, 160,
and 48 out of 273, in the order Sol, Terra, Luna, and GPT-5.5.
Terra and Luna therefore do not strictly follow their aggregate score
order at every tolerance. All model labels refer to the recorded
configurations; the experiment is not an external intelligence test.
All four configurations receive the same prompt, which explicitly asks
the agent not to finalize while budget remains. In that historical runner,
however, two consecutive replies without
a tool call caused the runner to finalize the session. The recorded
terminal status does not by itself identify
whether a particular stop was model-authored or runner-mediated. We therefore
treat the 11\% median budget use of GPT-5.5 as behavior of the complete
configuration.
In the four-agent participant set, paired problem-bootstrap score differences
are Sol minus Terra $0.054$ with interval $[-0.064,0.161]$, Sol minus Luna
$0.072$ with interval $[-0.038,0.167]$, and Sol minus GPT-5.5 $0.634$ with
interval $[0.537,0.698]$. Thus only the GPT-5.5 separation is stable under this
problem-resampling analysis; the three GPT-5.6 scores should not be read as a
statistically established total order. The corresponding ten-repeat campaign
and its terminal-status accounting are reported in
Appendix~\ref{app:ten-repeat}.

\section{Selected Optimization Histories}
\label{app:cases}
The examples below provide complementary views of the full experimental
results rather than substitutes for the population-level analyses.
Historical examples provide context and counterexamples, the audited trace
establishes provenance for the main-text case, and the standardized case
records document selected trajectories from the current campaign. Tables
report recorded observations, while figures summarize temporal or spatial
patterns. Clean objective values are used only where explicitly stated for
scoring or post-hoc visualization and are never exposed to the agent.

\paragraph{Historical exploration and screening.}
We inspected all ten repetitions of four preselected problems from the earlier
Claude Code rotated cohort: HAIRY, EGGCRATE,
ROSENBR, and HIMMELBG. This cohort is distinct from the current
built-in-runner primary comparison. These problems were selected for
interpretability and do not constitute a random sample. A read-only screen
identified cases in which an improvement followed at least five consecutive
finite evaluations without an incumbent decrease exceeding
$10^{-12}+10^{-8}|f_{\rm anchor}|$. The subsequent improvement was
additionally required to exceed $0.01\max(1,|f_{\rm best}|)$; these
thresholds are selection conventions, not statistical tests. All 40 traces
passed the integrity checks, with warnings for legacy trace fields. Ten traces
contained a flagged event, all inside trust-region campaigns. None of these
flags alone identifies a new model decision or establishes escape from a
local minimum. The remaining 30 traces contained no flagged event.
No objective evaluations or model calls were added.

On HAIRY repetition 5, the agent resumed a numerical campaign after reporting
stagnation; evaluation 17 reduced the best value from $49.57$ to $24.03$.
The reduction was computed inside the resumed numerical method. On EGGCRATE
repetition 0, a reset campaign and three subsequent design points did not
materially improve on the best value near $9.488$; the agent stopped after
39 evaluations. Together, the two cases show both a successful continuation
and an unsuccessful exploratory reset, providing context for interpreting
selected trajectories.

\paragraph{Interpretation boundary.}
A flat best-so-far curve indicates observed stagnation, not a certified local
minimum. A tool call followed by improvement establishes temporal
association, not that the tool call was necessary or sufficient for the
improvement. Recorded evaluations can arise from internal steps of a single
numerical campaign, so we link each evaluation to the corresponding campaign
and agent tool request before attributing a choice to the model.
We do not reconstruct missing reasoning from tool names. A
comparative case requires the same problem instance, transformed start,
budget, information policy, and declared controller configurations. Claims
that an alternative cannot escape require more than one unsuccessful run.

\begin{samepage}
	\paragraph{Audited trace: \textsc{SNAIL} repetition 09.}
	The main-text plot (Figure~\ref{fig:snail-case-main}) is reconstructed from a
	single recorded trajectory of GPT-5.6 Terra under Codex with medium
	reasoning on \textsc{SNAIL} repetition 9 in the ten-repeat perturbed-start
	study with noise. The recorded sequence is summarized below:

	\begin{center}
		\small
		\begin{tabular}{@{}>{\raggedright\arraybackslash}p{3.25cm}
				>{\raggedright\arraybackslash}p{8.35cm}@{}}
			\toprule
			Recorded transition & Line search (evaluations 78--80), followed by a Newton step (evaluation 81); these evaluations left the incumbent at 13.38.\\
			Switch marker & After evaluation 81, the agent issued a broad \code{run\_trust\_region} call with reset and radius factor 5.0, beginning with evaluation 82.\\
			Post-switch result & Evaluations 82--100; best recorded value $1.157\times10^{-3}$ at evaluation 98 within the 100-evaluation budget. The clean evaluator check at the same point is $2.101\times10^{-4}$.\\
			\bottomrule
		\end{tabular}
	\end{center}

	The clean objective values and contours in the figure are evaluator-side
	checks and were not available to the agent. The case therefore documents a
	recorded transition and its subsequent trajectory without establishing
	causality or certifying that the preceding incumbent was a local minimum.
\end{samepage}

\paragraph{Historical counterexample.}
In the earlier Claude Code rotated HIMMELBG trial at repetition 4, four
trust-region campaigns used 8, 10, 12, and 2 evaluations after the initial
point. Evaluation 33 reached $9.45\times10^{-22}$. A subsequent Newton step
produced an objective value of $0.0544$ at evaluation 34, after which the
environment retained the earlier incumbent. The agent then terminated the
session with 66 evaluations remaining. This case illustrates incumbent
preservation and the distinction between numerical-method termination and
session-level stopping, rather than a successful escape or a benefit from the
final tool switch.

\paragraph{Outcome-level historical summaries.}
Single-function histories illustrate the numerical scale of these
differences. On perturbed ROSENBR, NextEval's median clean history-best value
is $3.34\times10^{-14}$, compared with $3.37\times10^{-3}$ for COBYQA and
$5.92\times10^{-2}$ for NEWUOA. On CLIFF, the corresponding values are
$0.2009$ for NextEval and $0.1998$ for both model-based references.
ROSENBR shows a large difference in final numerical precision, whereas CLIFF
shows only a small difference in the final attained values.

\begin{figure}[ht]
	\centering
	\begin{tabular}{@{}cc@{}}
		\small ROSENBR: perturbed start & \small CLIFF: perturbed start\\
		\includegraphics[width=.48\linewidth]{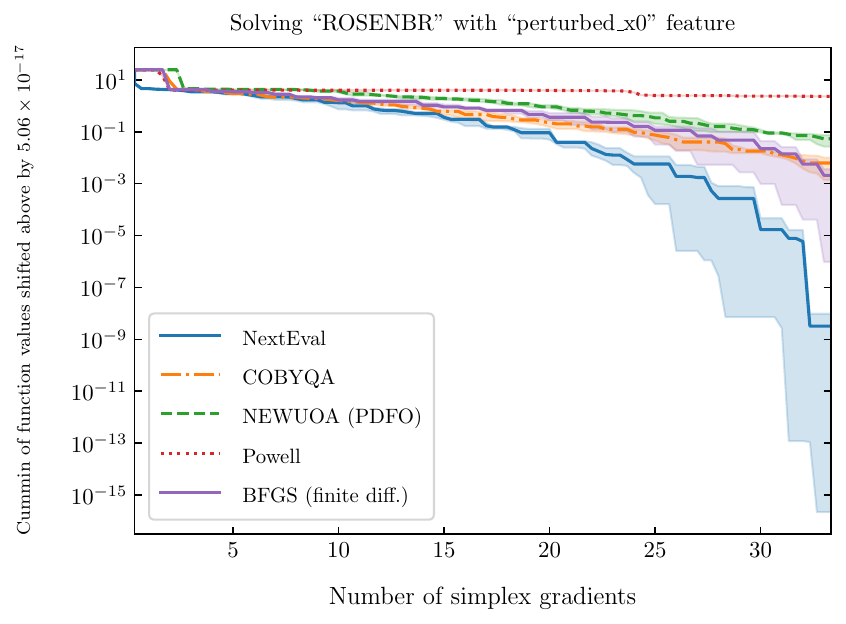}&
		\includegraphics[width=.48\linewidth]{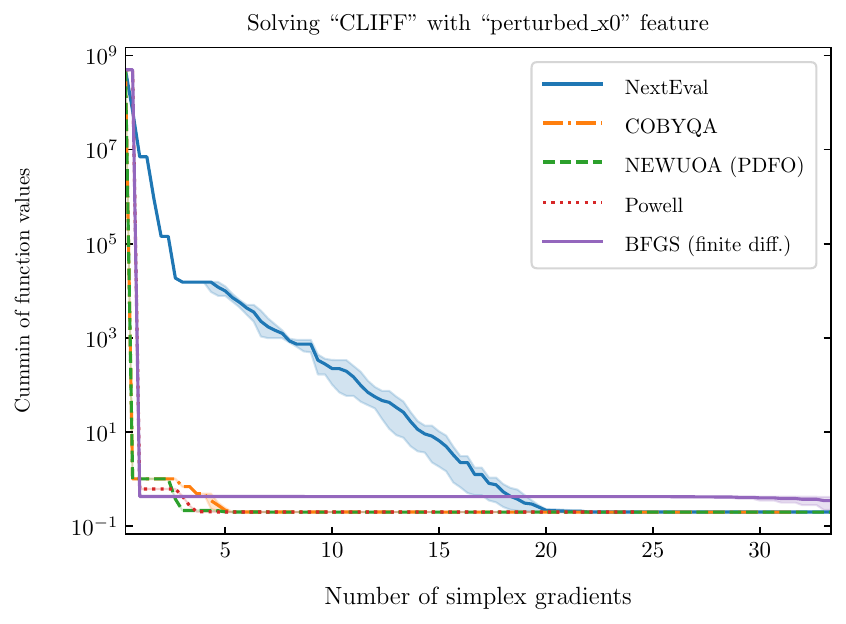}\\
		\small ROSENBR: perturbed start + noise & \small HIMMELBG: perturbed start + noise\\
		\includegraphics[width=.48\linewidth]{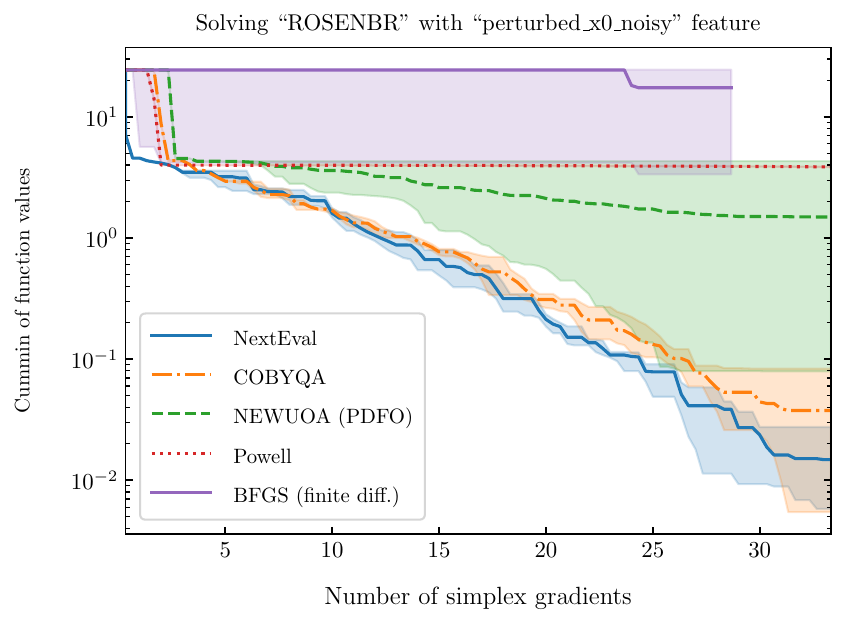}&
		\includegraphics[width=.48\linewidth]{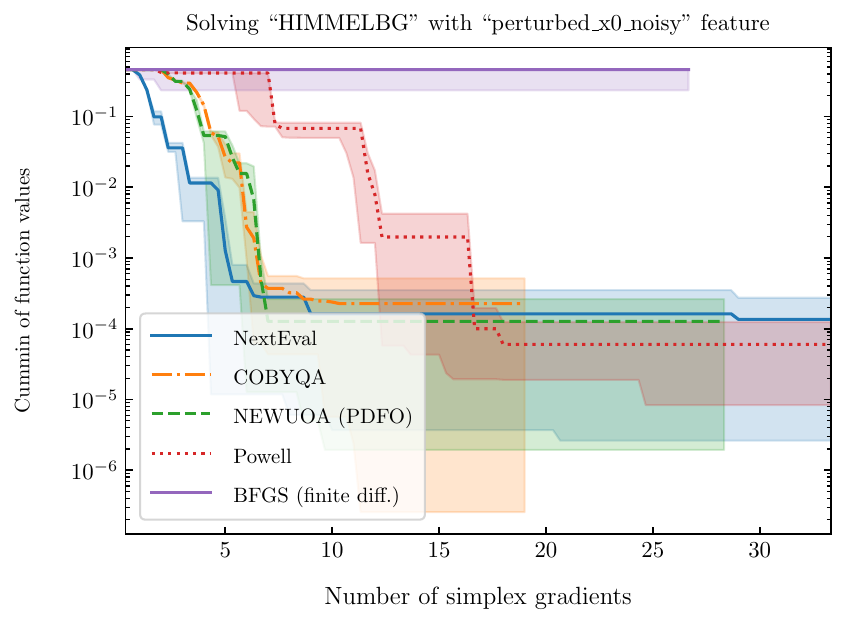}
	\end{tabular}
	\caption{NextEval--DFO histories for ROSENBR, CLIFF, and HIMMELBG under
		perturbed starts and, for the lower row, observation noise. All three
		repetitions are included in each archived plot. These histories describe
		outcomes rather than a tool-level causal ablation.}
\end{figure}

\begin{figure}[ht]
	\centering
	\includegraphics[width=.48\linewidth]{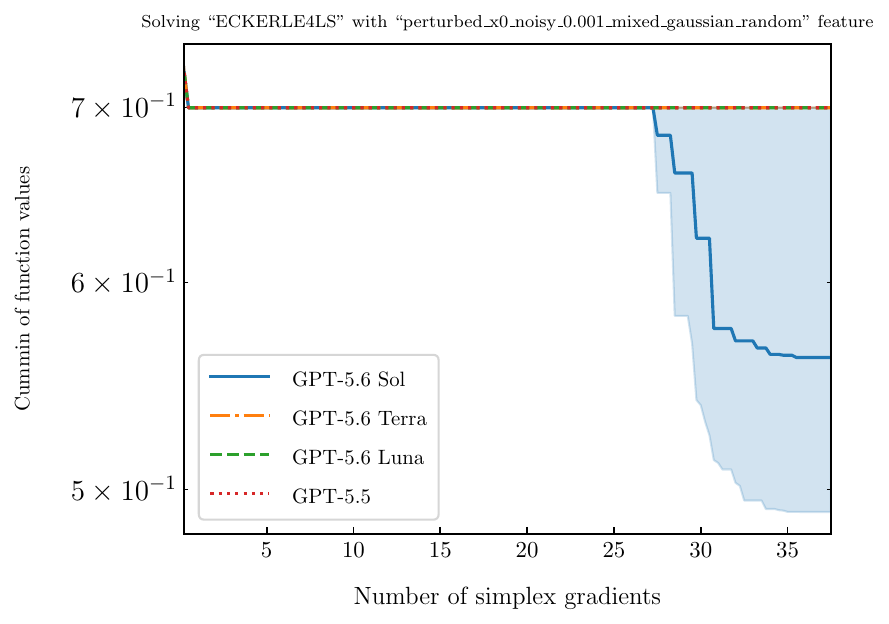}\hfill
	\includegraphics[width=.48\linewidth]{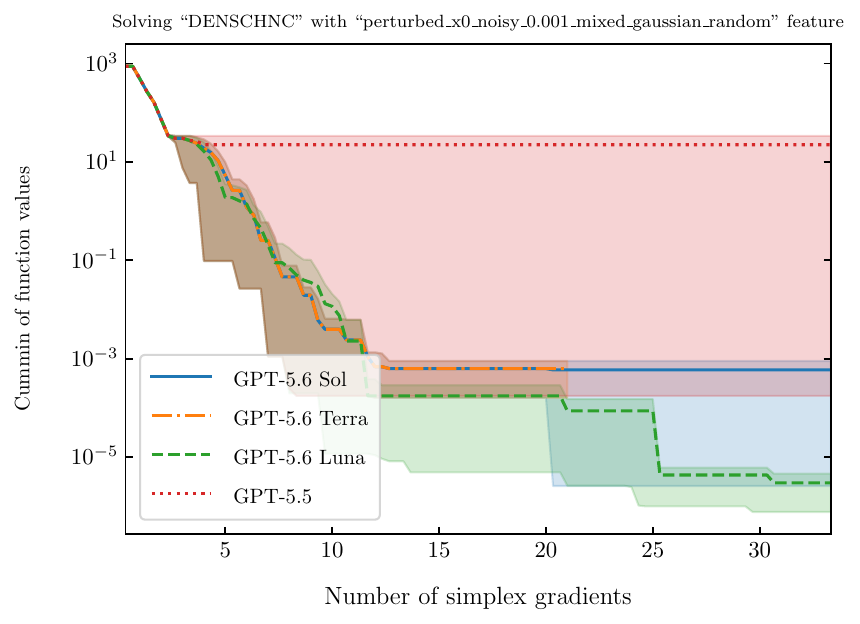}
	\caption{Two illustrative GPT histories from the earlier three-repeat
		campaign. They are not ten-repeat summaries; aggregate
		results are determined by the full comparison set rather than these selected
		examples.}
\end{figure}

\subsection{Standardized Ten-Repeat Case Records}
\label{app:trajectory-cases}
This subsection reports eight selected records from the ten-repeat campaign:
two paired ROSENBRTU outcome records and six traces presented in a common
format. The SNAIL repetition-09 row repeats the audited trace above so that
the comparison table uses the same fields for each case. For each record, we
list the problem, repetition, model configuration, evaluation budget,
recorded action sequence, observed objective values, and paired classical
solver observations. These traces are descriptive records rather than
population summaries.

\begin{figure*}[t]
	\centering
	\includegraphics[trim=0 0 0 36,clip,width=.98\textwidth]{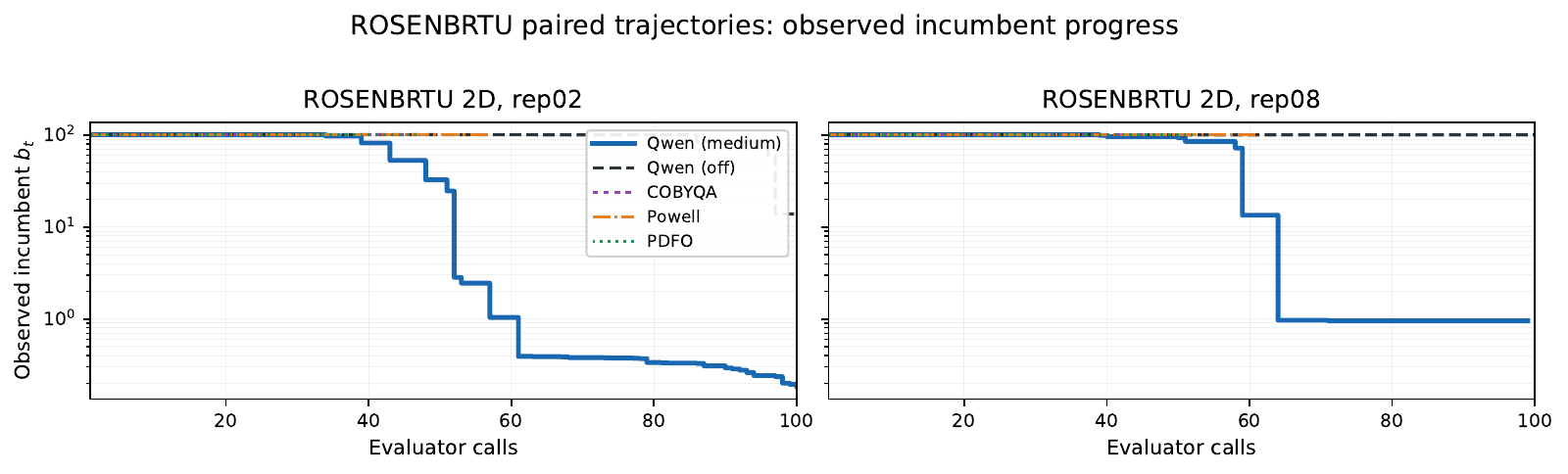}
	\caption{Observed incumbent curves for paired two-dimensional ROSENBRTU
		records under Qwen 3.8 Flash with thinking set to medium or off and the three
		paired classical references. Each curve is the cumulative minimum of the
		method's recorded noisy observations. The vertical axis reports the observed
		incumbent and the horizontal axis reports evaluator calls.}
	\label{fig:app-rosenbrtu-paired}
\end{figure*}

\paragraph{Paired ROSENBRTU outcome records.}
These two-dimensional records use Qwen 3.8 Flash through the Codex
whole-episode MCP harness, with thinking set to medium or off. Each repetition
uses a paired starting point, a perturbation strength of $0.001$, mixed
Gaussian observation noise with strength $0.001$, and a 100-evaluation
budget. The paired classical references are COBYQA, Powell, and PDFO; PDFO
uses UOBYQA for this problem. The numerical outcomes are summarized in
Table~\ref{tab:app-rosenbrtu-paired} and
Figure~\ref{fig:app-rosenbrtu-paired}.

\begin{table*}[t]
	\centering
	\small
	\caption{Observed incumbent values for paired two-dimensional ROSENBRTU
		records under the listed Qwen settings. The classical entry is the best
		observed value among COBYQA, Powell, and PDFO in the same repetition.}
	\label{tab:app-rosenbrtu-paired}
	\begin{tabularx}{\textwidth}{@{}>{\centering\arraybackslash}p{0.12\textwidth}
			>{\centering\arraybackslash}p{0.13\textwidth}
			>{\centering\arraybackslash}p{0.13\textwidth}
			>{\centering\arraybackslash}p{0.14\textwidth}
			>{\centering\arraybackslash}p{0.18\textwidth}X@{}}
		\toprule
		Repetition & Qwen medium & Qwen off & Best classical & Best classical value & Medium strict improvements\\
		\midrule
		rep02 & $0.1843$ & $13.88$ & PDFO & $100.7$ & 35\\
		rep08 & $0.9525$ & $100.6$ & PDFO & $100.8$ & 13\\
		\bottomrule
	\end{tabularx}
\end{table*}

\paragraph{Action-level two-dimensional traces.}
Table~\ref{tab:app-gpt10-cases} records three two-dimensional traces with
100-evaluation budgets and paired Powell, COBYQA, and PDFO observations.

\begin{table*}[t]
	\centering
	\scriptsize
	\caption{Recorded observations for three action-level two-dimensional
		ten-repeat traces. All objective values are observed values; the final column
		reports paired classical observations and reported termination conditions.}
	\label{tab:app-gpt10-cases}
	\begin{tabularx}{\textwidth}{@{}p{0.17\textwidth}p{0.42\textwidth}X@{}}
		\toprule
		Case & Model and recorded evaluation sequence & Paired classical observations\\
		\midrule
		ROSENBRTU rep04 & GPT-5.5; evaluations 53--58; incumbent $100.8$ before three line-search calls without a strict decrease; \code{scale\_restart} followed, with observed values $92.82$ (evaluation 58), $3.642$ (61), and $0.8773$ (89). & Powell: 96/100 calls after its last strict improvement, then evaluation-limit termination at $100.7$; COBYQA/PDFO: 43/45 and 46/48 plateau calls before a trust-region radius lower bound.\\
		SNAIL rep09 & GPT-5.6 Terra; evaluations 78--82; see the audited case above for the full recorded action sequence; observed value $1.157\times10^{-3}$ at evaluation 98. & Powell: successful termination after 73 calls at $14.23$, with 50 calls after its last strict improvement; COBYQA/PDFO: 11/52 and 7/59 plateau calls, with PDFO near $13.93$.\\
		LOGHAIRY rep08 & GPT-5.6 Terra; evaluations 38--39; incumbent $0.4815$ and a Newton candidate of $0.8568$; subsequent trust-region calls reached $0.1805$ at evaluation 67. & Powell: 72/100 calls after its last strict improvement and evaluation-limit termination at $6.539$; COBYQA/PDFO: 32/49 and 10/48 plateau calls before a trust-region radius lower bound, with best observed value near $6.529$.\\
		\bottomrule
	\end{tabularx}
\end{table*}

\begin{figure*}[t]
	\centering
	\includegraphics[width=.98\textwidth]{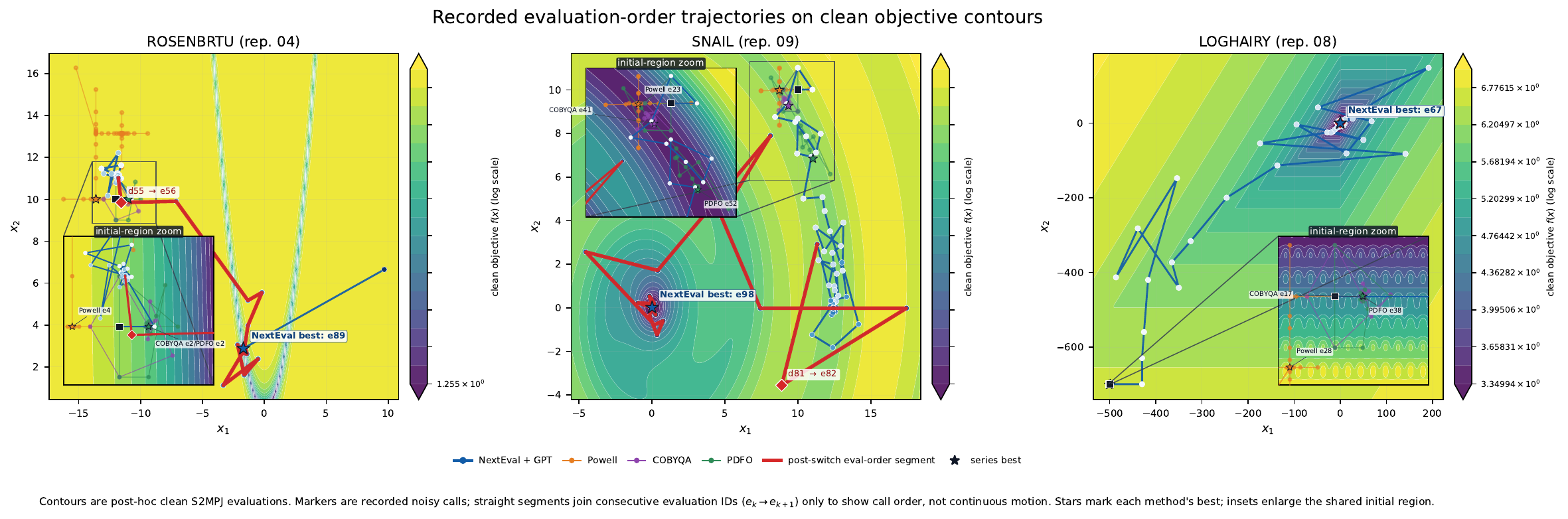}
	\caption{Recorded decision trajectories for three two-dimensional ten-repeat
		traces. Contour colors show post-hoc clean S2MPJ objective evaluations on a
		logarithmic scale. Blue, orange, purple, and green paths and markers show
		NextEval+GPT, Powell, COBYQA, and PDFO, respectively. Each straight connector
		joins consecutive recorded evaluator calls and does not denote continuous
		optimizer motion. Red overlays highlight the segment from the evaluation
		immediately preceding the selected action change to the evaluation attaining
		the selected NextEval best. Stars mark each method's best recorded
		observation, and insets enlarge the shared initial regions. The clean contours
		were not available to the agent.}
	\label{fig:app-gpt10-trajectory}
\end{figure*}

The evaluated points and recorded action changes for these three traces are
shown in Figure~\ref{fig:app-gpt10-trajectory}.

\paragraph{Supplementary dimensional traces.}
The remaining three traces use the same recorded-history format; their
settings and outcomes are summarized in
Table~\ref{tab:app-gpt10-supplement}.

\begin{table*}[t]
	\centering
	\scriptsize
	\caption{Recorded observations for three supplementary ten-repeat traces. The
		dimension and budget columns identify the experimental setting; the final
		column reports paired classical observations.}
	\label{tab:app-gpt10-supplement}
	\begin{tabularx}{\textwidth}{@{}p{0.17\textwidth}c r r p{0.30\textwidth}X@{}}
		\toprule
		Case & Dim. & Budget & Final observed & Recorded evaluation sequence & Paired classical observations\\
		\midrule
		SNAIL rep08 & 2 & 100 & $-2.191\times10^{-3}$ & Line search without a strict decrease near $13.93$ $\rightarrow$ broad trust region to $1.987\times10^{-3}$ $\rightarrow$ Newton to $6.75\times10^{-4}$ $\rightarrow$ trust-region continuation. & Powell/\allowbreak COBYQA/\allowbreak PDFO best values $14.24/\allowbreak 14.14/\allowbreak 13.98$; plateau tails 23/100, 3/51, and 17/55 calls.\\
		ECKERLE4LS rep08 & 3 & 150 & $2.06\times10^{-4}$ & Coordinate search without a strict decrease $\rightarrow$ trust-region restart to $0.4459$ $\rightarrow$ line search to $0.4198$ $\rightarrow$ trust-region refinement. & Powell/\allowbreak COBYQA/\allowbreak PDFO best values $0.6974/\allowbreak 0.6979/\allowbreak 0.6973$; plateau tails 61/141, 35/66, and 46/75 calls.\\
		MGH09LS rep03 & 4 & 200 & $-3.94\times10^{-4}$ & Line search and broad trust-region calls without a strict decrease $\rightarrow$ coordinate search. & Powell/\allowbreak COBYQA/\allowbreak PDFO best values $2.627\times10^{-3}/\allowbreak 3.200\times10^{-3}/\allowbreak 1.231\times10^{-2}$; plateau tails 8/200, 29/98, and 27/145 calls.\\
		\bottomrule
	\end{tabularx}
\end{table*}

The archived agent, COBYQA, and PDFO histories were not rerun. Powell was
evaluated using the same frozen start, feature seed, noisy evaluator, and
per-case budget. Before the Powell runs were added, the reconstruction
reproduced all archived COBYQA observations exactly. No model call was added.

\section{Controlled Agent-Configuration Study Details}

\subsection{Ten-Repeat Model and Reasoning Studies}
\label{app:ten-repeat}

\begin{figure}[ht]
\centering
\includegraphics[width=.47\linewidth]{figures/gpt-performance.pdf}\hfill
\includegraphics[width=.47\linewidth]{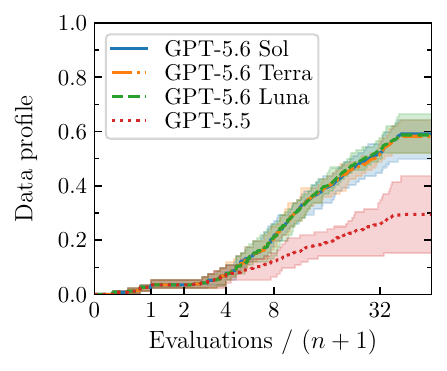}
\caption{GPT model comparison with Codex and medium reasoning fixed.
History-performance and history-data profiles at $10^{-6}$ use all 92
problems and ten repetitions. Targets are computed within the four models;
shading shows the repetition range, not a confidence interval.}
\label{fig:gpt-profiles}
\end{figure}

\paragraph{GPT model-choice study.}
Holding Codex and medium reasoning fixed, we compare GPT-5.6 Sol, Terra,
Luna, and GPT-5.5 under perturbed starts plus noise. Their aggregate scores
are 0.997, 0.993, 0.981, and 0.571. At $10^{-6}$ they attain 544, 536, 540,
and 271 targets out of 920, respectively (Figure~\ref{fig:gpt-profiles}).
For MISRA1ALS, three start perturbation seeds yielding nonfinite values
are replaced by finite-start reserve seeds before solver execution;
the noise seeds are unchanged.

The four-model comparison contains $92\times4\times10=3{,}680$ agent
episodes. Its original rows are unchanged; 40 MISRA1ALS episodes
were added under the recorded host and CLI deviation above. Four historical
GPT-5.5 episodes timed out, but their available numerical histories are
retained: DENSCHNA (8), BRKMCC (9),
ROSENBRTU (5), and KOWOSB (5).

All four GPT configurations receive the same instruction not to finalize
while budget remains. The runner can nevertheless convert repeated replies
without tool calls into a terminal status, so stopping behavior is a property
of the complete model--harness configuration. Median budget use is 100\%
for the GPT-5.6 configurations and 13\% for GPT-5.5. The available numerical
histories from timed-out sessions remain in the profile comparison.
Appendix~\ref{app:models} retains the earlier three-repeat snapshot separately.

\paragraph{Qwen reasoning-setting study.}
The Qwen final snapshot contains $92\times5\times10=4{,}600$ cells,
including 1,840 agent episodes, from 4,950 attempts. The archive retains
the final authorized terminal attempt for each slot rather than selecting
the attempt with the best objective value, and completed parents are not
retried. Across the archive, 350 failed historical attempts remain in the
audit trail. The selected-attempt profiles therefore measure objective-
evaluation efficiency rather than recovery overhead. Failures mix transport
and execution causes and are not all attributed to model quality.
MISRA1ALS remains in this snapshot; three COBYQA final outputs are nonfinite,
and the archive preserves the scorer's nonfinite convention.

For Qwen 3.8 Flash under Codex, the history-performance area scores are
0.924 with thinking off and 1.000 with medium reasoning. Medium obtains the
lower median final clean value on 53 problems, compared with 39 for thinking
off. At the individual repetition level, medium records 484 lower final values,
426 higher values, and 10 ties. These within-problem repetitions are not 920
independent problem families.

\begin{figure}[ht]
\centering
\includegraphics[width=.47\linewidth]{figures/qwen-performance.pdf}\hfill
\includegraphics[width=.47\linewidth]{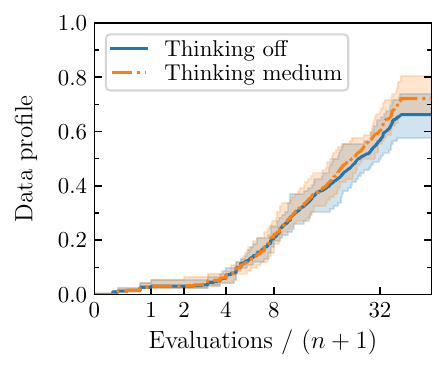}
\caption{Ten-repeat Qwen reasoning-setting comparison. Both configurations
use NextEval with Codex and Qwen 3.8 Flash; only the reasoning setting
(off versus medium) changes. OptiProfiler history-performance and data
profiles are shown at $10^{-6}$ over all 92 problems. References are
recomputed within these two agents, and shading shows the repetition range
rather than a confidence interval.}
\end{figure}

Label changes were checked to leave numerical profile scores unchanged.
Profile generation reloads
the saved OptiProfiler data without rerunning the solvers. These snapshots do
not reuse the earlier cohort's bootstrap intervals.

\subsection{DeepSeek Execution-Harness Comparison}
\label{app:harness-study}
Both participants use the same numerical implementation, DeepSeek V4 Flash with thinking
off, the same NextEval instructions and 18 tools, and ten matched repeats
under perturbed starts plus noise. The provider is shared within each pair.
Both receive a 32,000-token output cap, temperature zero, noise-informed
instructions, and no evaluation cache. The harness changes: the built-in
runner uses its digest-based context management, whereas Claude Code
retains its own wrapper, context management, and recovery. This compares
complete harnesses, not one internal mechanism.
The 92-problem comparison adds 20 MISRA1ALS episodes on macOS to the original
Linux archive, retaining its existing histories. The supplement uses the
finite-start reserve seeds described above, with unchanged noise seeds;
three transport-interrupted attempts are preserved alongside their replacements.

\begin{figure}[ht]
\centering
\includegraphics[width=.47\linewidth]{figures/harness-performance.pdf}\hfill
\includegraphics[width=.47\linewidth]{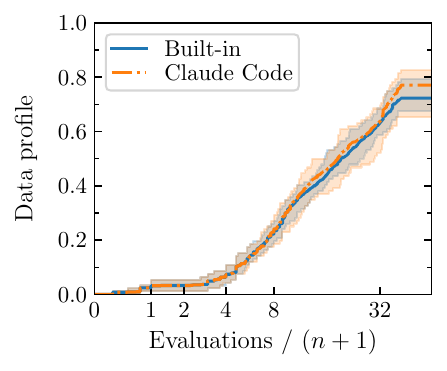}
\caption{DeepSeek V4 Flash with the built-in runner
and Claude Code. OptiProfiler history-performance
and data profiles at $10^{-6}$ use all 92 problems with ten repeats.
Targets are computed within these two configurations; shading shows
the repetition range.}
\label{fig:harness-profiles}
\end{figure}

\begin{table}[ht]
\centering\small
\caption{History-based profile scores, averaged over tolerances
$10^{-1}$ through $10^{-10}$. Higher is better.}
\label{tab:harness-scores}
\begin{tabular}{@{}lrr@{}}
\toprule
Harness & Performance & Data\\
\midrule
Built-in & 0.947 & 0.966\\
Claude Code & 0.999 & 1.000\\
\bottomrule
\end{tabular}
\end{table}

All recorded histories are retained, including partial built-in sessions.
One authorized rerun of a runtime error instead ended at the output-token
cap; substituting its history leaves the reported profile scores unchanged.
The archive preserves both attempts and all transport retries. Claude Code
performs better in this configuration; this does not identify which harness
component causes the difference.

\paragraph{Earlier Qwen comparison.}
The historical three-repeat Qwen 3.8 Flash study compared Codex with direct
tool calling with a common numerical implementation in the two arms.
Its 273 complete pairs yielded
136/8/129 lower/equal/higher final values for Codex. Different instruction
wrappers and API transports make it a separate study; it is not pooled
with the matched DeepSeek comparison.

\clearpage
\section{Planned Mechanism Tests}
\label{app:ablation-design}
This section specifies follow-up controls; it contains no ablation results.
The solver is the combination of controller, numerical tools, and their
execution environment. We do not claim that gains from this combination
are exclusively due to the language model.

\paragraph{Replace the controller, retain the environment.}
A non-LLM controller can select the same tools from the same permitted
history. A deterministic controller may still adapt to observations;
stochasticity is not the distinction between an agent and a numerical rule.
A simple first reference is to continue the implemented trust-region method
under a declared budget and stopping rule, without changing its numerical
implementation. This tests whether method composition improves on a strong
component, not whether an LLM dominates every possible controller. A learned
router or reinforcement-learned policy is a separate research direction,
not a prerequisite for the present lightweight solver.

\paragraph{Retain the model, reduce numerical support.}
The complementary comparison holds model, inference setting, and session
management fixed, while exposing only a budgeted point-evaluation interface
and raw permitted history instead of the numerical toolset and summaries.
Even this minimal interface must retain the oracle boundary, ledger, budget,
and error handling. It is therefore not literally ``no harness.'' If general
code execution remains available, it must be declared: the agent could
implement its own numerical methods. A narrower alternative removes only
history-analysis tools or only continuation-state reuse.

\paragraph{Comparison and interpretation.}
These are two controlled substitutions, not automatically an orthogonal
model-present/harness-present factorial experiment. Removing both leaves no
defined sampling policy until an additional baseline is specified. Freeze
controls on development problems, use matched task/feature seeds and budgets,
and report all repetitions, failures, stopping behavior, and model costs.
Do not choose the strongest single method retrospectively on the test set.
Case studies should include improvement, neutral, and unsuccessful sequences.
An observed plateau alone cannot establish a local minimum; a claimed
advantage from a particular action requires matched comparisons, or repeated
continuations from equivalent states and remaining budgets.

\section{Numerical and Execution Details}
\label{app:implementation}
\paragraph{Quadratic models.}
In normalized coordinates, the trust-region tool fits
$q(d)=c+g^\top d+\tfrac12d^\top Hd$ through selected observations,
using the least-Frobenius-change problem
\begin{equation}
 \min_{c,g,H=H^\top}\tfrac12\|H-H_{\rm prev}\|_F^2
 \quad\text{subject to}\quad q(d_i)=y_i.
\end{equation}
The linear terms are unpenalized. Geometry checks and regularization
support numerical stability. A Newton jump uses the fitted model and
its definiteness and validity checks before evaluating a candidate.
The implementation supplies these classical numerical mechanisms;
the agent controls their use and evaluation allowances.

\paragraph{Termination and ownership.}
Global budget exhaustion, explicit finalization, and operational
interruptions are distinct from a local method's stopping condition.
Stopping outcomes are interpreted according to the rules used in each
experiment. Final results and partial histories must be retained when a
session is interrupted.
In-process runners are trusted code. API ownership by itself does not
prove isolation against malicious Python introspection or same-user
filesystem access. The strict benchmark executor must enforce its
oracle, filesystem, and network boundaries independently.

\section{Design Rationale and Related Approaches}
\label{app:design-rationale}
\paragraph{Solution prediction and adaptive decisions.}
By a one-shot solution mapping, we mean a predictor that returns a solution
from a problem description or fixed observations without requesting further
objective evaluations. Such predictors can be useful when the problem
family has shared structure. Our general-purpose oracle setting does not
assume that this structure is known or that a new objective follows a
training distribution. In the absence of further restrictions, different
objectives can agree on a finite set of observations while having different
minimizers. Predictions from those observations alone cannot distinguish
them. Additional evaluations can provide new information, but a finite-budget
adaptive policy also has no guarantee of success on arbitrary objectives.
The choice of an adaptive policy is therefore a design principle, not an
impossibility result for learned optimization or a distribution-free guarantee
for NextEval. It also does not rule out learning the policy itself.

\paragraph{Why use an agent?}
General-purpose agents already combine tool use with observation-dependent
decisions \citep{yao2023react,merrill2026terminalbench}, and reusable agent
frameworks support persistent execution and context
\citep{langgraph2026overview}. These capabilities make the proposed solver
practical to implement; they do not establish that a particular model will
make good DFO decisions. Our hypothesis concerns using these capabilities
to control numerical methods, not substituting LLM-generated arithmetic
for numerical computation. Direct evaluations of LLM-based optimization
report numerical limitations \citep{rw20260912_huang_llm_bbo}, but the present
experiments do not isolate whether numerical tools, retained state, or
agent decisions account for the performance differences.

\paragraph{Adjacent uses of agents.}
NextEval controls numerical methods during an individual solve. Program
search and heuristic generation instead develop algorithms across tasks,
as in FunSearch and ReEvo
\citep{rw20260906_romera_funsearch,rw20260906_ye_reevo}.
Agents also support explicit mathematical modeling and application-specific
optimization \citep{rw20260906_ahmaditeshnizi_optimus,rw20260906_jiang_optagent_buildings}.
These are complementary directions, not methods ruled out by our choice
of a scalar objective interface.

\clearpage
\section{Numerical Tool Catalog}
\label{app:tool-catalog}
The tools in Table~\ref{tab:tools} implement the numerical operations used
by the agent in Section~\ref{sec:tools}. Each method retains its own state;
analysis tools use existing observations without new objective calls.

\begin{table}[htbp]
\centering\small
\caption{The numerical action space. Method state is separate across
families. The final two rows use existing observations only.}
\label{tab:tools}
\begin{tabular}{@{}p{3.1cm}p{5.0cm}p{4.1cm}@{}}
\toprule
Operation & Numerical work & Retained information\\
\midrule
Trust-region campaign & Quadratic interpolation, step acceptance, geometry repair & Radius, resolution, sample set, Hessian\\
Line-search campaign & Bracketing and safeguarded parabolic search & Direction and successful scale\\
Poll / coordinate search & Directional tests and step-size adaptation & Poll length / one scale per axis\\
Newton jump & Safeguarded minimizer of a fitted quadratic & Model checks and recent outcome\\
Geometry rebuild & Construct a usable local sample set & Samples and geometry report\\
\midrule
Model preview & Fit a surrogate and inspect its candidate & No new objective evaluations\\
Situation / history & Progress, scales, readiness, past decisions & No new objective evaluations\\
\bottomrule
\end{tabular}
\end{table}
\end{document}